%% file: main.tex
\documentclass[10pt,twocolumn,letterpaper]{article}

\usepackage[pagenumbers]{cvpr} 

\input{preamble}

\definecolor{cvprblue}{rgb}{0.21,0.49,0.74}
\usepackage[pagebackref,breaklinks,colorlinks,citecolor=cvprblue]{hyperref}

\usepackage[dvipsnames]{xcolor}
\usepackage{soul,xcolor}
\usepackage{lipsum}
\usepackage{colortbl, tabulary, etoolbox}

\usepackage{algorithm}
\usepackage{algpseudocode}

\floatname{algorithm}{Algorithm}

    \usepackage{amsmath}
    \usepackage{amsmath,amssymb}
    
    \DeclareMathOperator{\EX}{\mathbb{E}}
\usepackage{pythonhighlight}
\usepackage{xr}
\makeatletter
\newcommand*{\addFileDependency}[1]{
\typeout{(#1)}
\@addtofilelist{#1}
\IfFileExists{#1}{}{\typeout{No file #1.}}
}\makeatother

\newcommand*{\myexternaldocument}[1]{%
\externaldocument{#1}%
\addFileDependency{#1.tex}%
\addFileDependency{#1.aux}%
}
\myexternaldocument{main_supp}

\author{Alexander Vilesov$^{\star}$
\and
Pradyumna Chari$^{\star}$
\and 
Achuta Kadambi
\and
University of California, Los Angeles\\
\tt\small\{vilesov, pradyumnac\}@ucla.edu, achuta@ee.ucla.edu \\
}

\begin{document}
\newcommand{\paperAcronym}{CG3D}
\title{\paperAcronym: \textcolor{red}{C}ompositional \textcolor{red}{G}eneration for Text-to-\textcolor{red}{3D} \\ via  Gaussian Splatting}

\twocolumn[{%
\renewcommand\twocolumn[1][]{#1}%
\maketitle
    \captionsetup{type=figure}
    \vspace{-1cm}
    \begin{subfigure}[c]{\textwidth}
        \centering
        \includegraphics[width=\linewidth]{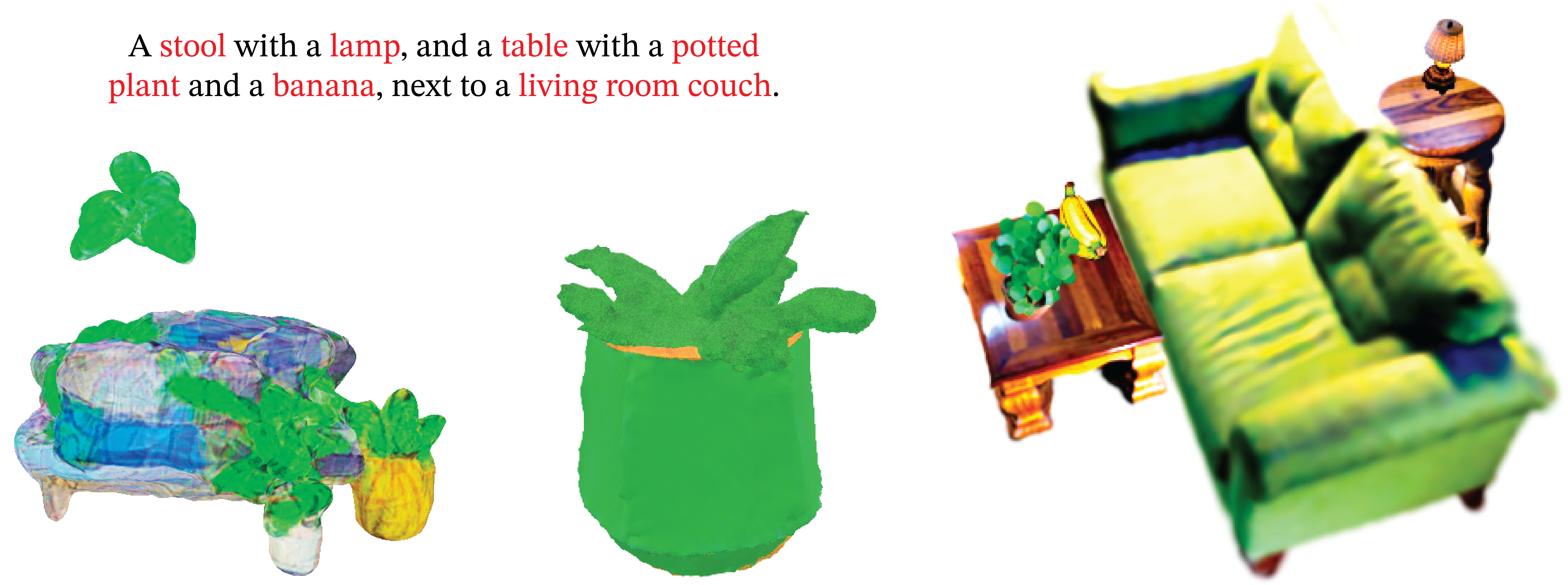}
        \label{fig:teaser}
    \end{subfigure}

    \vspace{-0.2cm}
    
    \begin{subfigure}[c]{0.31\textwidth}
        \centering    \caption{\footnotesize{\textsc{DreamGaussian~\cite{tang2023dreamgaussian}}}}
    \end{subfigure}
    \begin{subfigure}[c]{0.27\textwidth}
        \centering    \caption{\footnotesize{\textsc{Fantasia3D~\cite{chen2023fantasia3d}}}}
    \end{subfigure}
    \begin{subfigure}[c]{0.41\textwidth}
         \centering    \caption{\footnotesize{\textsc{Ours}}}
    \end{subfigure}
    \vspace{-8mm}
    
    \hfill\caption{\textbf{We propose a method for scalable, composable 3D generation from text prompts only.} Prior methods are unable to generate scenes consistent with detailed text, while the proposed method leverages explicit representations to enable 
 physically correct compositions, without any additional changes to the guiding diffusion model. 
    }
    \vspace{0.25cm}
    \label{fig:teaser}
}]
\setstcolor{red}

\renewcommand*{\thefootnote}{$\star$}
\setcounter{footnote}{1}
\footnotetext{Indicates equal contribution.}
\renewcommand*{\thefootnote}{\arabic{footnote}}
\setcounter{footnote}{0}

\input{sections/0_abstract}    
\input{sections/1_intro}

\input{sections/2_relatedworks}
\input{sections/3_method}

\input{sections/4_experiments}
\input{sections/5_limitations}
\input{sections/6_conclusion}

{
    \small
    \bibliographystyle{ieeenat_fullname}
    \bibliography{main}
}

\input{sections/A_suppl}

\end{document}

%% file: preamble.tex
\usepackage[dvipsnames]{xcolor}


%% file: sections/0_abstract.tex
\begin{abstract}
\vspace{-0.46cm}
With the onset of diffusion-based generative models and their ability to generate text-conditioned images, content generation has received a massive invigoration. Recently, these models have been shown to provide useful guidance for the generation of 3D graphics assets. However, existing work in text-conditioned 3D generation faces fundamental constraints: (i) inability to generate detailed, multi-object scenes, (ii) inability to textually control multi-object configurations, and (iii) physically realistic scene composition. In this work, we propose \paperAcronym, a method for compositionally generating scalable 3D assets that resolves these constraints. We find that explicit Gaussian radiance fields, parameterized to allow for compositions of objects, possess the capability to enable semantically and physically consistent scenes. By utilizing a guidance framework built around this explicit representation, we show state of the art results, capable of even exceeding the guiding diffusion model in terms of object combinations and physics accuracy. Project webpage: \href{https://asvilesov.github.io/CG3D/}{https://asvilesov.github.io/CG3D/} 
\end{abstract}

%% file: sections/1_intro.tex
\section{Introduction}
\label{sec:intro}

Current generative text-to-3D methods are incapable of producing scene-level results. While such methods have achieved remarkable performance at the object-level, scene-level prompts result in only parts of the scene being generated or complete failure to semantically adhere to the prompt. 
In this work we propose compositional scene generation from text by harnessing explicit 3D representations. Explicit representations decouple objects from the scene thus allowing users high flexibility for composition or editing at the object and scene level. 

We build upon a large body of existing work to achieve this. With the arrival of image diffusion models trained on large scale datasets, users have been able to create a variety of content. The control over the generation process now encompasses a range of inputs as a form of conditioning which include text~\cite{dhariwal2021diffusion, ho2022classifier}, images~\cite{zhang2023sine}, edge outlines~\cite{zhang2023adding}, and masks~\cite{avrahami2023blended}. The generative field is rapidly evolving with the aims of expanding to image reconstruction~\cite{feng2023score}, video~\cite{wu2023tune, blattmann2023align}, audio~\cite{kong2020diffwave}, and 3D objects~\cite{poole2022dreamfusion, chen2023fantasia3d, lin2023magic3d}. 3D object synthesis is a complex task which requires the fabricated object to be consistent with a given text prompt from any view. Several pioneering works~\cite{sanghi2022clip, mohammad2022clip, jain2022zero} attempted this through various forms of CLIP guidance. This was advanced by \citet{poole2022dreamfusion} through guiding the formation of Neural Radiance Field (NeRF) with a pre-trained image diffusion model. Subsequent works have focused on improving quality, diversity of objects, and extending to other 3D formats. However, few efforts toward text-to-3D generation have addressed compositionality. We aim to fill this gap by introducing a composition-based method that leverages explicit 3D representations. We summarize our major contributions below:
\begin{enumerate}
    \item A framework for compositional generation of scalable scenes using explicit radiance fields
    \item Physically realistic scene generation while maintaining object separability and fidelity
    \item A method to estimate object composition parameters (rotation, translation, scale) through score distillation sampling, without the need for object bounding boxes
\end{enumerate}

%% file: sections/2_relatedworks.tex
\section{Related Works}
\label{sec:related_works}

\subsection{3D Scene Representation}
A differentiable representation for a 3D scene is key to represent a wide range of scenes and to enable subsequent tasks such as text-to-3D generation.
A common method in recent years has been Neural Radiance Fields (NeRFs)~\cite{mildenhall2021nerf} which represent 3D scenes with a coordinate-based network that can be easily queried for novel view synthesis. Subsequent NeRF-based works have aimed at improving reconstruction quality~\cite{barron2021mip, verbin2022ref, sitzmann2020implicit, lindell2022bacon}, speed of training~\cite{muller2022instant}, representing large-scale scenes~\cite{tancik2022block}, and achieving reconstruction under constrained conditions~\cite{tiwary2023orca, ahn2023neural}. Recently, \citet{kerbl20233d} proposed an explicit representation for novel-view synthesis, that uses splatting~\cite{zwicker2001surface} of 3D anisotropic Gaussians to improve rendering speed. As opposed to these works, our focus is on \textit{generating} multi-object compositional scenes from text prompts, rather than representing existing scenes.

\subsection{Text-guided generation using diffusion}
Recent progress in image generation has been spear-headed by diffusion models~\cite{ho2020denoising}. Such models have been adapted to take in complex prompts and generate high-quality images and scenes closely aligning to the given prompt~\cite{rombach2022high, ho2022classifier, ruiz2023dreambooth, avrahami2023blended}. \citet{song2020denoising} devised a new sampling method to accelerate sampling through denoising diffusion implicit models. The generation of high-resolution images is achieved through a cascade of super-resolution models~\cite{saharia2022photorealistic, balaji2022ediffi} or performing the diffusion process in a low-resolution latent space where the resulting latents are decoded to image space~\cite{rombach2022high}. Text guidance was originally achieved through an explicit classifier model~\cite{dhariwal2021diffusion} and later through classifier free guidance (CFG)~\cite{ho2022classifier}. More elaborate guidance schemes have been devised to include additional inputs such as sketches and poses~\cite{zhang2023adding} for more control.  \\
With such progress in image generation, interest has surged in text-to-3D generation. Earlier works such as CLIP-forge~\cite{sanghi2022clip}, DreamField~\cite{jain2022zero}, and CLIP-mesh~\cite{mohammad2022clip} optimized their 3D models by maximizing text-image alignments scores. However, diffusion guidance was adapted to generate more detailed 3D primitives through the use of Score Distillation Sampling (SDS)~\cite{poole2022dreamfusion}. Ensuing work like Magic3D~\cite{lin2023magic3d} has improved qualitative results by adopting two-stage training where a course NeRF is first learned and then converted to a DMTET representation for further optimization, Fantasia3D~\cite{chen2023fantasia3d} decouples geometry and material using physically-based rendering to create high-quality meshes, and ProlificDreamer~\cite{wang2023prolificdreamer} introduced variational score distillation to improve upon the low-diversity of samples generated with SDS. We would like to highlight concurrent work (on ArXiv) focused on 3D asset generation using the 3D Gaussian Splatting framework that aim to achieve fast generation of 3D assets~\cite{tang2023dreamgaussian} and high quality 3D assets by incorporating 3D diffusion priors~\cite{chen2023text,yi2023gaussiandreamer} . Compared to all these prior methods, our focus is on generating physically realistic compositional 3D scenes from text input only, while allowing for individual object control. 

\subsection{Compositional 3D generation}
Compositional 3D generation entails generating a cohesive scene from several lower-level primitives. For the purposes of this paper, lower-level primitives will be objects. One of the first steps to scene synthesis is defining the spatial relationships between objects which is typically done with scene graphs~\cite{chang2014learning, qi2018human, zhu2018modeling}. Traditionally, one of the main applications of scene synthesis has been room layouts, with early approaches constraining the problem using interior design rules and object frequency distributions~\cite{merrell2011interactive, yeh2012synthesizing}. Newer methods adopt the generative machine learning paradigm using VAEs~\cite{purkait2020sg, yang2021scene}, GANs~\cite{yang2021indoor}, and Diffusion models~\cite{tang2023diffuscene}, where scene priors are learned instead of handcrafted such that scene generation models are automatic and end-to-end. Other works have focused on composition through decomposition~\citet{yang2021learning, niemeyer2021giraffe}. Several works have begun to explore the usage of pre-trained image diffusion models to expand the range of compositions~\cite{lin2023componerf, Po2023Compositional3S, Cheng2023Progressive3d} by constraining geometry for each object with a user-defined bounding box. Our focus is on generating scalable scenes in a realistic manner, from text input only without bounding-boxs or training a diffusion model.

%% file: sections/3_method.tex
\begin{figure}
    \centering
    \includegraphics[width=1\linewidth]{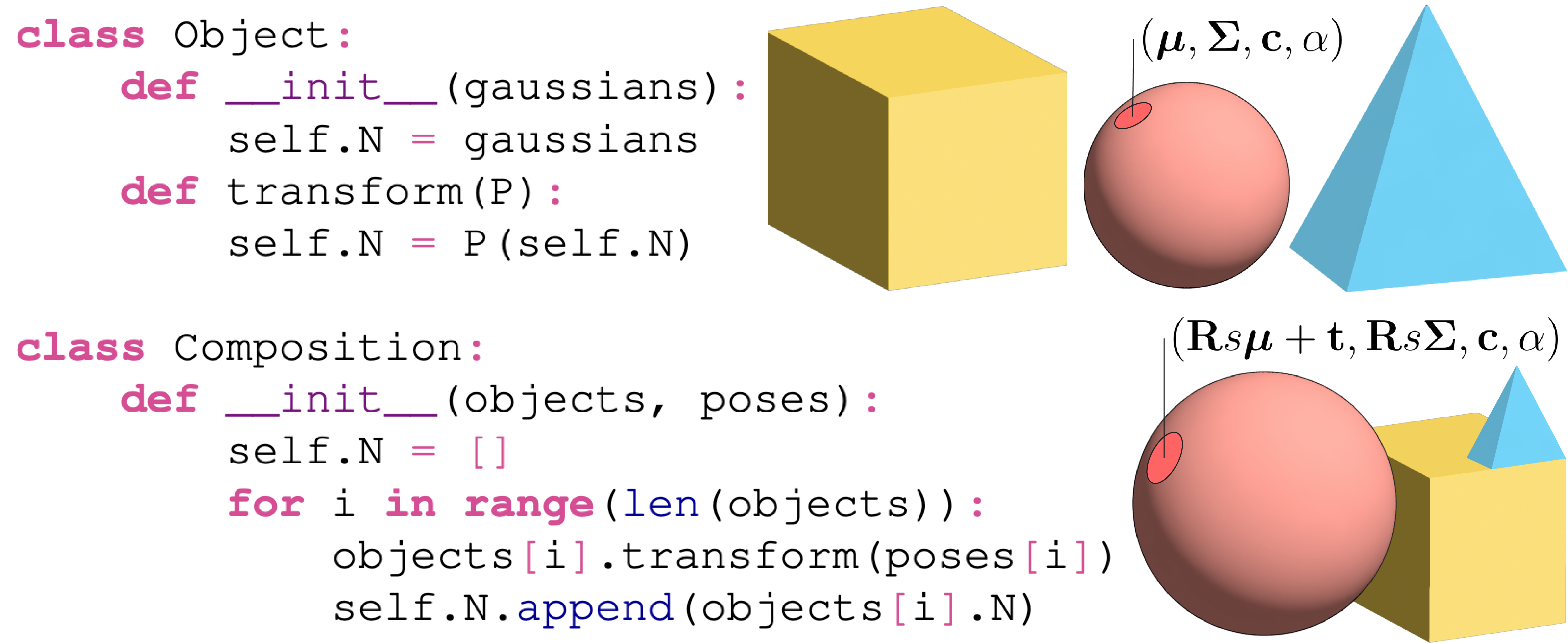}
    \caption{\textbf{We realize multi-object scenes through a Gaussian radiance field.} Pseudocode to enable compositionality in Gaussian radiance fields incorporating rotation, translation, and scale to convert 3D Gaussians from object to composition coordinates.}
    \label{fig:gaussians_code}
\end{figure}

\section{Compositional generation}

Our goal is to generate a compositional 3D scene $\mathcal{S}$ given a text prompt $\mathbf{y}$, and access to a 2D image diffusion model. We consider a general scene $\mathcal{S}$ consisting of a set of $K$ objects, $\mathcal{O}=\{\mathbf{O}_i \ | \ i \in [1,...,K]\}$.
Each object in $\mathcal{O}$ is represented by a set of Gaussians, $\theta$. An image, $\mathbf{X}$, can then be generated by rendering the set of 3D Gaussians, $\mathbf{X}=\mathcal{R}\{\theta\}$, using Gaussian splatting~\cite{kerbl20233d}. The value of a pixel at location $\mathbf{p}$ is found by projecting each Gaussian onto the image plane and alpha-blending their contributions~\cite{zwicker2001surface}:
\begin{equation}
\begin{aligned}
    \mathcal{R}\left\{ \mathbf{p}, \mathbf{M}, \mathcal{\theta} \right\} = \sum_{i \in \mathcal{\theta}} \mathbf{c}_i \alpha_i \prod_{j=1}^{i-1} (1-\alpha_j), \\ 
    \text{with } \ \alpha_i = o_i e^{-\frac{1}{2}(\mathbf{p}-\underline{\boldsymbol{\mu}_i})^{T} \underline{\mathbf{\Sigma}}^{-1} (\mathbf{p}-\underline{\boldsymbol{\mu}_i}))},
\end{aligned}
\end{equation}

where $\mathbf{c}_i$, $o_i$, $\boldsymbol{\mu}_i$, and $\mathbf{\Sigma}_i$ represent the color, opacity, position, and covariance, respectively, of the $i$th 3D Gaussian. Additionally, $\underline{\boldsymbol{\mu}_i}$ and $\underline{\mathbf{\Sigma}_i}$ represent the 2D projected counterparts of the mean and covariance, in screen-space coordinates. 

To render a compositional scene, we require a transformation from object to composition coordinates, consisting of a rotation $\mathbf{R}$, translation $\mathbf{t}$, and scale $s$. These transformations apply to the individual Gaussian means and variances. We therefore define the scene to also contain the interactions between objects, $\mathcal{P}=\{\mathbf{P}_{i,j}  \ | \ \forall \ i,j \in [1,...,K],\ i\neq j\}$, such that $\mathcal{S}=\{\mathcal{O},\mathcal{P}\}$. Here, $\mathbf{P}_{i,j}=(\mathbf{R}_{i,j},\mathbf{t}_{i,j},s_{i,j})$.  \cref{fig:gaussians_code} illustrates this radiance field structure and shows how this representation combines multiple objects into one data structure. 

\begin{figure}[b]
    \centering
    \includegraphics[width=0.9\linewidth]{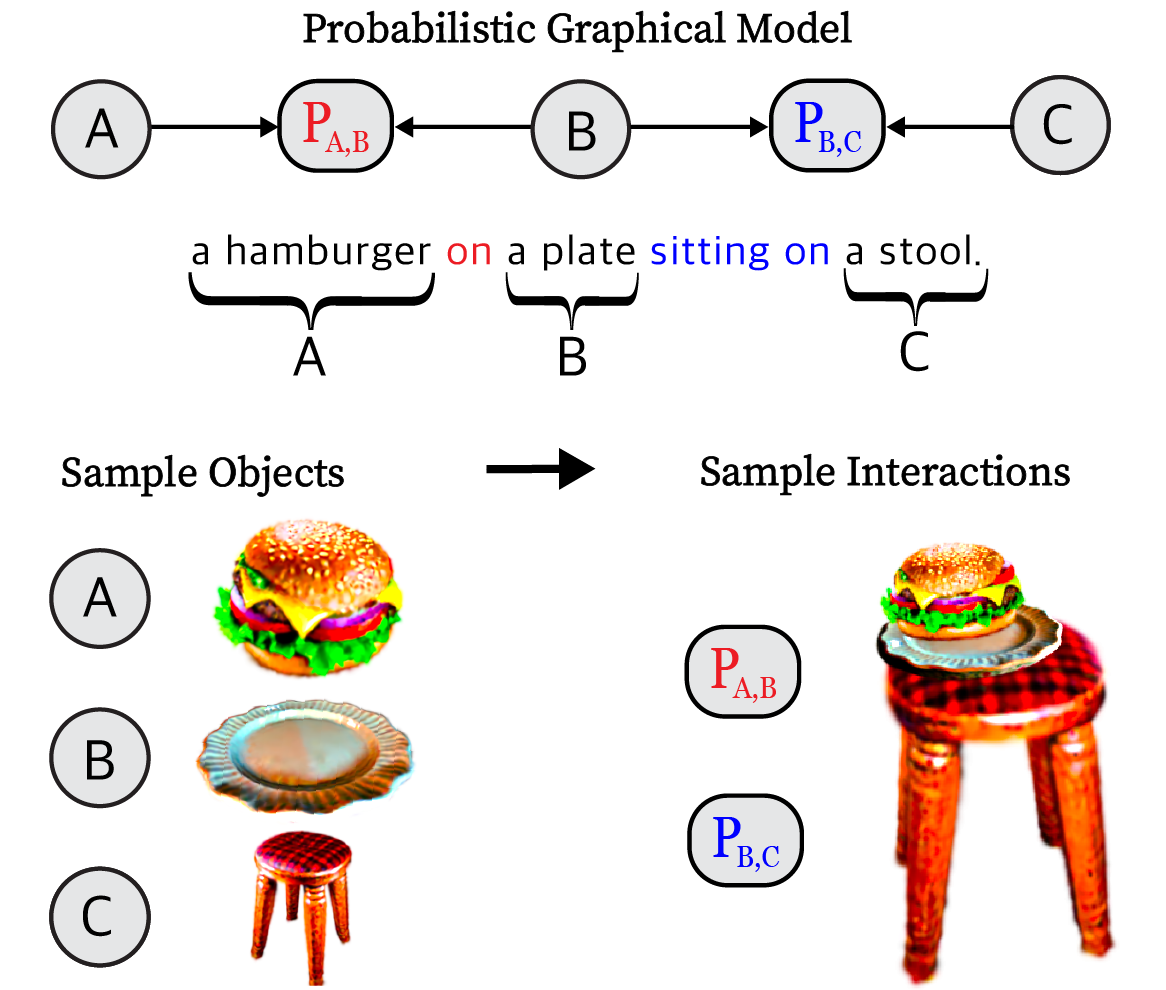}
    \caption{\textbf{Our method achieves compositional generation through ancestral sampling of a PGM of the scene.} We first sample objects followed by their pairwise interactions.}
    \label{fig:pipeline}
\end{figure}

Such an explicit compositional representation is key to enable scalable scene generation. The input to our method is the text prompt $\mathbf{y}$ manually deconstructed by the user into a scene graph. The scene graph's set of object nodes and interaction nodes are textual descriptions of the objects and interactions between them, respectively, thus forming a one-to-one correspondence to $\mathcal{O}$ and $\mathcal{P}$ in $\mathcal{S}$. That is, $\mathbf{y}$ is represented by,
\begin{equation}
    \mathbf{y}=\{\mathbf{y}_k | k\in [1,\dots,K]\}\, \cup \{\mathbf{y}_{i,j} | i,j\textrm{ s.t. }\mathbf{P}_{i,j}\in \mathcal{P}\},
\end{equation}
where $\mathbf{y}_k$ is the textual description for object $\mathbf{O}_k$, and $\mathbf{y}_{i,j}$ is the textual description of the interaction between objects $\mathbf{O}_i$ and $\mathbf{O}_j$.

We wish to maximize the probability of a compositional scene given a text prompt,
\begin{equation}
    p(\mathcal{R} \{ \mathcal{S} \}|\mathbf{y})=p(\mathcal{R} \{ \mathcal{O},\mathcal{P} \}|\mathbf{y}).
    \label{eq:comp_step_1}
\end{equation}
$\mathcal{R}\{\cdot\}$ denotes the rendering operator, which we will drop subsequently for notational convenience. This joint probability can be decomposed as follows:
\begin{equation}
    p(\mathcal{O},\mathcal{P}|\mathbf{y}) = p(\mathcal{O}|\mathbf{y})\cdot p(\mathcal{P}|\mathcal{O},\mathbf{y}).
    \label{eq:comp_step_2}
\end{equation}
The textual scene graph can then be interpreted as a probabilistic graphical model (PGM) where the directed edges are transformed such that the tails are at the object nodes and the heads are at the interaction nodes as illustrated in \cref{fig:pipeline}. With this formulation we make two assumptions: (a) the objects in $\mathcal{O}$ are independent of each other, and (b) the interaction between two objects $\mathbf{P}_{i,j}$ are conditionally dependent only on their connected objects $\mathbf{O}_i$ and $\mathbf{O}_j$. These assumptions arise out of the inability of image diffusion models (which we use for guidance) to generate and represent scenes with a large number of objects. Therefore our main formulation is as follows:
\begin{multline}
    p(\mathcal{S}|\mathbf{y}) = \prod_{k=1}^K \underbrace{p(\mathbf{O}_k  | \mathbf{y}_{k})}_{\text{\cref{sec:object_generation}}} \cdot \\
    \prod_{i,j\textrm{ s.t. }\mathbf{P}_{i,j}\in \mathcal{P}}\underbrace{p(\mathbf{R}_{i,j}, \mathbf{t}_{i,j}, s_{i,j} | \mathbf{y}_{i,j}, \mathbf{O}_i, \mathbf{O}_j)}_{\text{\cref{sec:interaction_params}}}.
\end{multline}

Through this formulation, we can generate a scene through ancestral sampling, by first generating the objects (\cref{sec:object_generation}) followed by their interactions (\cref{sec:interaction_params}). We note that such a formulation creates limitations that constrain generation abilities: (i) inability to represent interactions that require object geometry changes, (ii) explicit knowledge of all objects and interactions, and (iii) intersection events that may occur between two objects on separate branches of the graph. However, as our results show, the expressivity of this representation remains high, and able to cover a range of scene configurations.
\label{sec:comp_generation}

\subsection{Interaction Parameter Estimation}
\label{sec:interaction_params}
\begin{figure}[t]
  \centering
   \includegraphics[width=0.9\linewidth]{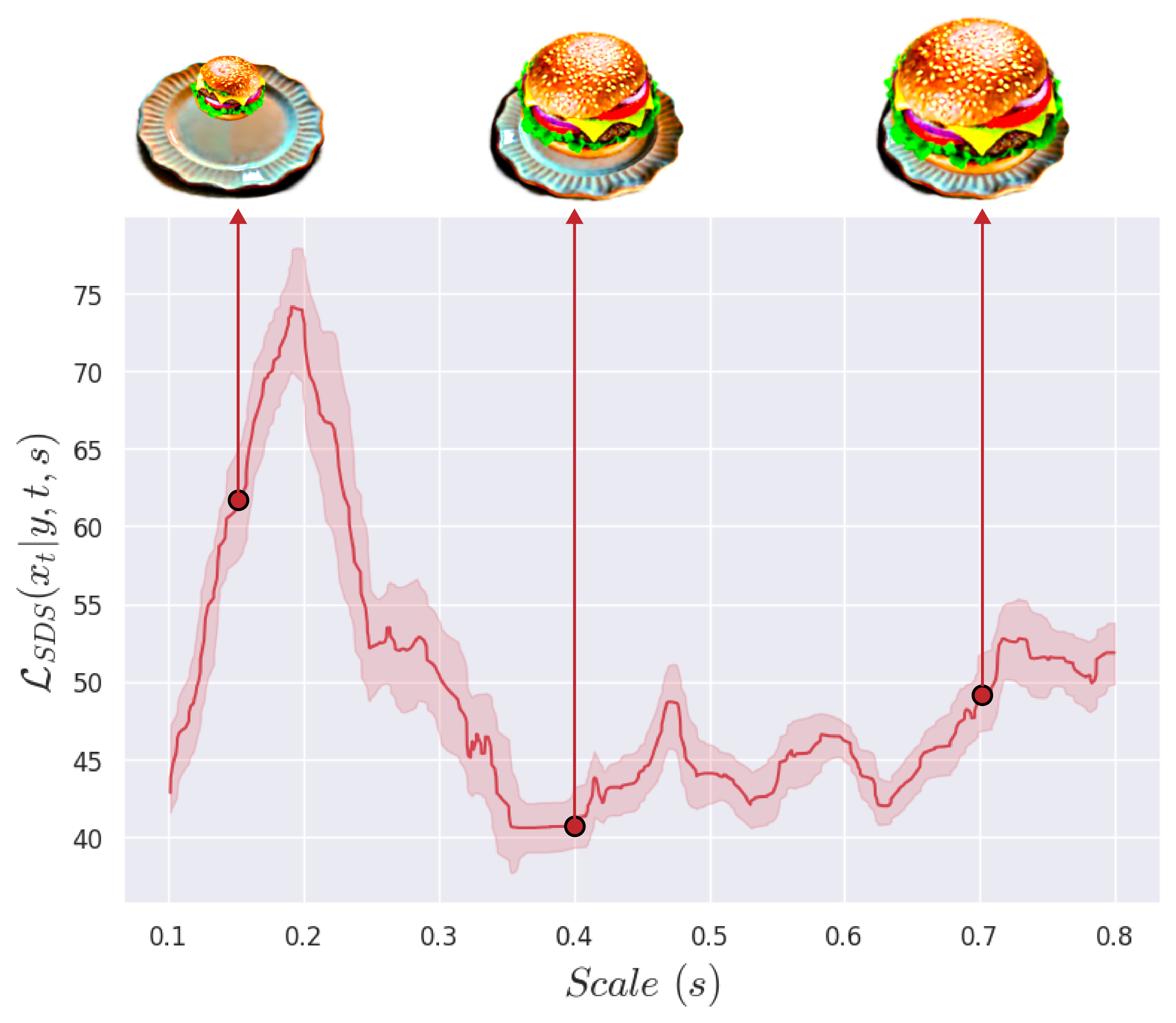}

   \caption{\textbf{Gradient descent optimization is poorly conditioned for estimating optimal} $\mathbf{R}_{2,1}$, $s_{2,1}$ \textbf{and} $\mathbf{t}_{2,1}$. Here, we show an anomaly in the SDS loss for unnaturally small $s_{2,1}$. Similar anomalies exist in the estimation of $\mathbf{t}_{2,1}$.}
   \label{fig:pose_scale_init}
\vspace{-4mm}
\end{figure}

Without loss of generality, let us consider a two-object scene with objects $\mathbf{O}_1$ and $\mathbf{O}_2$. We set  $\mathbf{O}_1$ to be the anchor object, with respect to which the interaction parameters for $\mathbf{O}_2$, ($\mathbf{R}_{2,1}$,$\mathbf{t}_{2,1}$, $s_{2,1}$), are defined. The anchor object, by definition, does not move. Our goal is to sample the interaction parameters such that,
\begin{multline}
    (\mathbf{R}^*_{2,1},\mathbf{t}^*_{2,1},s^*_{2,1})=\\
    \underset{\mathbf{R}_{2,1},\mathbf{t}_{2,1},s_{2,1}}{\arg \max} p(\mathbf{R}_{2,1}, \mathbf{t}_{2,1}, s_{2,1} | \mathbf{y}_{2,1}, \mathbf{O}_1, \mathbf{O}_2).
\end{multline}
\subsubsection{Score Distillation Sampling for Interaction Parameters}
\label{sec:sds_interaction}
We wish to infer configuration parameters consistent with text guidance $\mathbf{y}_{2,1}$. We begin by using Score Distillation Sampling (SDS) to estimate these parameters. SDS provides a gradient update to guide generation through:
\begin{equation}
    \nabla_{\boldsymbol{\theta}} \mathcal{L}_{SDS}(\boldsymbol{\theta}) = \EX_{t,\epsilon} \left[w(t)(\epsilon_{\psi}(\mathbf{X}_t | \mathbf{y},t) - \epsilon)\frac{\partial \mathbf{X}}{\partial \boldsymbol{\theta}} \right],
\end{equation}
where $\epsilon_{\psi}$ is the denoising function of the diffusion model $\psi$ that predicts sampled noise $\epsilon$ applied to an image $X_t$ at a particular time-step $t$ and text-prompt $\mathbf{y}$.
Deriving intuition from~\citet{poole2022dreamfusion}, we view $\mathcal{L}_{SDS}$ as an estimate of the likelihood of an image scene. However, we interpret it as a function of $\mathbf{R}_{2,1}, \mathbf{t}_{2,1}, s_{2,1}$ and denote this function as $\mathcal{F}$:
\begin{equation}
    (\mathbf{R}^*_{2,1},\mathbf{t}^*_{2,1},s^*_{2,1})=\underset{\mathbf{R}_{2,1},\mathbf{t}_{2,1},s_{2,1}}{\arg \min} \mathcal{F}.
    \label{eq:config_sds_eq}
\end{equation}
We refer to $\mathcal{F}$, as a function of interaction parameters, to be the configurational liklihood function (CLF), since it indicates the viability of a particular scene configuration.

Optimizing \cref{eq:config_sds_eq} through stochastic gradient descent is a natural first attempt. However this is non-trivial: the CLF is found to be extremely noisy, and with specific behavioral quirks that make the configuration generation task especially difficult. Figure~\ref{fig:pose_scale_init} shows the CLF for the prompt ``A hamburger on a plate", varied across various scales of the hamburger. The following details may be inferred:
\begin{enumerate}
    \item CLF provides an extremely noisy loss landscape leading to potentially getting stuck in local optima
    \item There are multiple solutions (local minima) in terms of semantic viability
    \item CLF provides inaccurate guidance at lower scales
\end{enumerate} 
A particularly unusual aspect is the inaccurate guidance at lower scales $s_{2,1}$. Specifically, we note that below a certain value of $s_{2,1}$, $\gamma$, the loss function value progressively decreases, leading to false optima at lower scales. This behavior is consistent across multiple compositions. 

We observe a thematically similar behavior affecting the estimation of the translation $\mathbf{t}_{2,1}$. Configurations that involve $\mathbf{O}_2$ being occluded by $\mathbf{O}_1$ from the perspective of the camera viewpoints (such as in a plate under a table) lead to lower CLF values than configurations without such occlusions (even though these configurations may violate the provided text conditioning). While we defer a more detailed discussion of this behavior of the SDS to the supplement, we note that these anomalies pose a significant challenge to estimating configuration parameters. Without appropriately accounting for these effects, SDS-based sampling of $(\mathbf{R}_{2,1},\mathbf{t}_{2,1},s_{2,1})$ would unnaturally prefer small objects and in occluded configurations.

It is therefore impractical and inaccurate to apply gradient descent to solve \cref{eq:config_sds_eq}. We propose a different approach: in the $s_{2,1},\mathbf{t}_{2,1}$ space, we perform random Monte Carlo (MC) sampling to cover a greater representative span of the CLF landscape, and provide ourselves a better chance of arriving at global optima. We use an alternating optimization approach, as follows:

\begin{enumerate}
    \item First, a joint Monte Carlo search in the $s_{2,1},\mathbf{t}_{2,1}$ space
    \item Freeze $s_{2,1}$, sample and search for $\mathbf{t}_{2,1}$
    \item Freeze $\mathbf{t}_{2,1}$, sample and search for $s_{2,1}$
    \item Repeat steps 2 and 3 $L=3$ times while keeping $\mathbf{R}_{2,1}$ fixed
\end{enumerate}

\begin{figure}[t]
  \centering
   \includegraphics[width=0.32\linewidth]{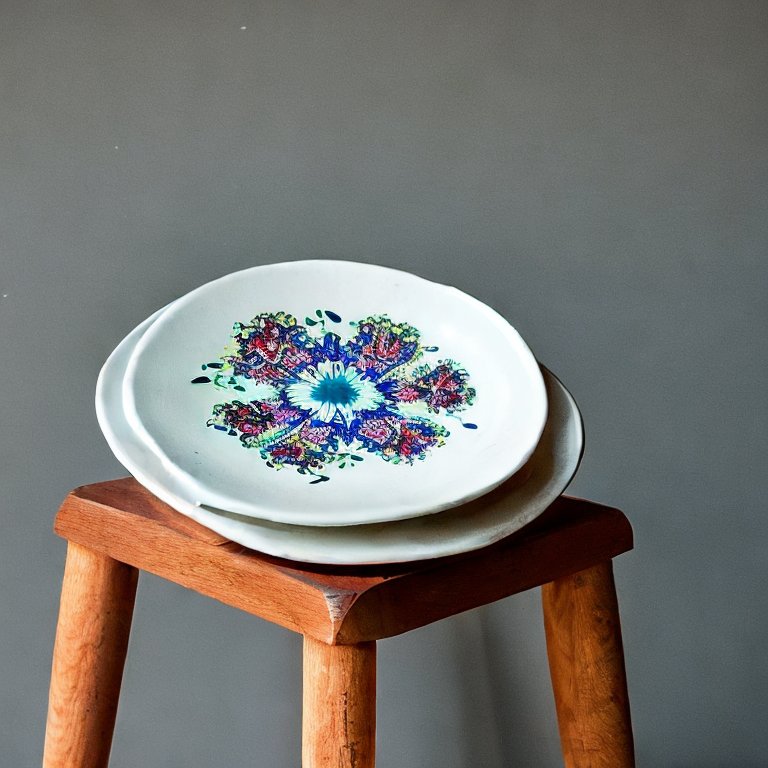}
   \includegraphics[width=0.32\linewidth]{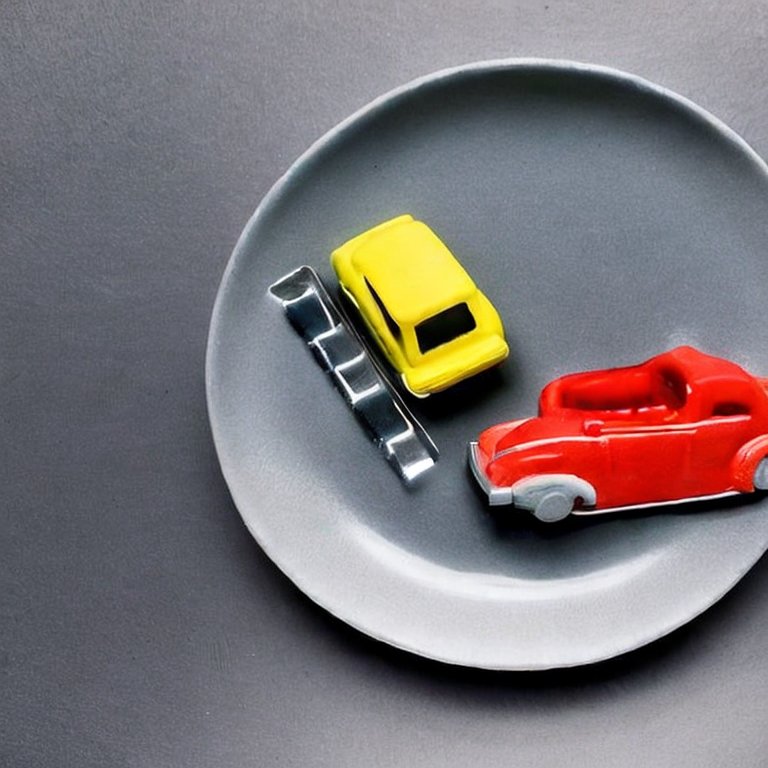}
   \includegraphics[width=0.32\linewidth]{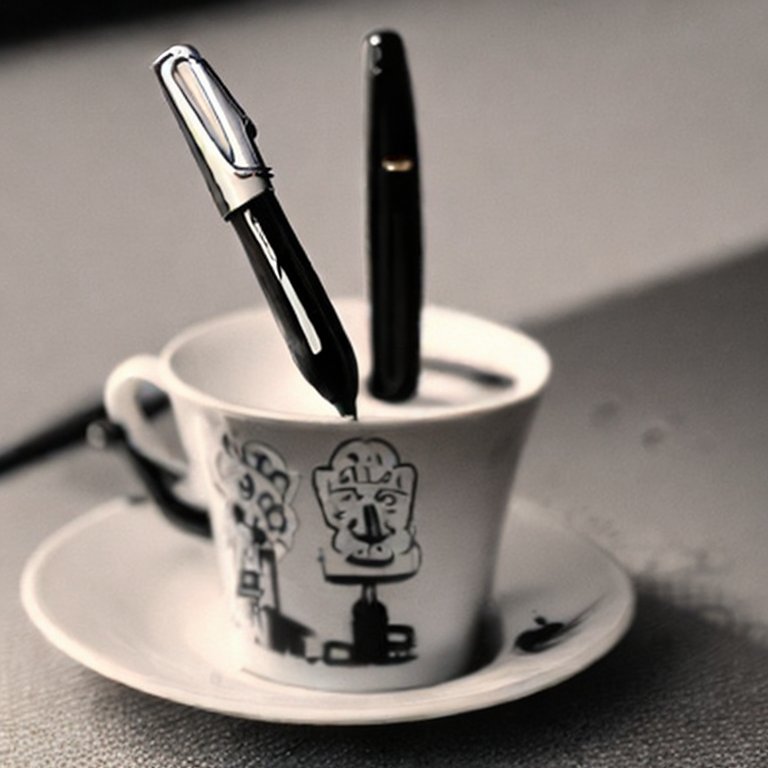}
    \caption{\textbf{Diffusion models, such as Stable Diffusion v2.1~\cite{rombach2022high} are unable to always adhere to physical laws such as gravity, even for image generation.} Additional physical guidance is required for realistic-looking scene compositions.}
   \label{fig:physics_accuracy}
   \vspace{-4mm}
   
\end{figure}

The proposed MC sampling accounts for the noisy loss landscape of the CLF, but is still susceptible to the scale and translation anomalies. To \textbf{correct for the scale anomaly}, we sample scale above the anomaly threshold $\gamma$, which we estimate. To ensure this does not limit our ability to handle scenes with smaller scales, we also enable camera radius reduction across steps of the CLF MC sampling, if the current $s_{2,1}$ estimate is close to the threshold.

To \textbf{accommodate the occlusion CLF anomaly}, we scale down the CLF $\mathcal{F}$ with a visibility function $v(s_{2,1},\mathbf{t}_{2,1})$, when estimating $\mathbf{t}_{2,1}$. The function $v(\cdot)$ is designed to be very low in the case of occlusions and high otherwise, and is realized using the viewspace gradients of our compositional Gaussian field. A detailed description of our initialization technique may be found in the supplement.

\subsubsection{Enabling Physically Accurate Composition} 
\label{sec:phy_losses}

Optimizing for \cref{eq:config_sds_eq} provides an estimate for $\mathbf{R}^*_{2,1},\mathbf{t}^*_{2,1},s^*_{2,1}$. However, this estimate is still limited by the prior of the underlying base diffusion model. Most significant among these is the lack of explicit physical constraints. This manifests in the form of unrealistic image generation that ignores physics laws such as gravity and contact forces (Figure~\ref{fig:physics_accuracy}). 
For a 3D generation and configuration task such as ours, this leads to the estimated $\mathbf{P}_{2,1}$, being physically unrealistic.

Leveraging the explicit nature of our Gaussian radiance field, we propose a two-stage estimation: the first being guided by SDS (\cref{sec:sds_interaction}), and the second being a finetuning based on physics-enforcing constraints $\mathcal{L}_{\textrm{Phys}}$, along with SDS. Note that gradient descent on SDS can provide meaningful guidance at this stage, post the initialization. That is,
\begin{equation}
\begin{split}
\mathbf{P}'_{2,1}=(\mathbf{R}'_{2,1},\mathbf{t}'_{2,1},&s'_{2,1})=\underbrace{\underset{\mathbf{R}_{2,1},\mathbf{t}_{2,1},s_{2,1}}{\arg \min} \mathcal{F}}_{\text{SDS init. (\cref{sec:sds_interaction})}},\\
(\mathbf{R}^*_{2,1},\mathbf{t}^*_{2,1},s^*_{2,1})=&\underbrace{\underset{\mathbf{R}_{2,1},\mathbf{t}_{2,1},s_{2,1}}{\arg \min} \mathcal{F}+\mathcal{L}_\text{Phys}\big\vert_{\textrm{init. at }\mathbf{P}'_{2,1}}}_{\text{Phys. finetune}}.
\label{eq:config_sds_eq_v3}
\end{split}
\end{equation}
$\mathcal{L}_{\textrm{Phys}}$ models two phenomena: gravity and normal contact forces. Practically, at this step, we freeze the scale parameter $s_{2,1}$, and only optimize for the rotation and translation, for greater training stability.

Consider object $\mathbf{O}_1$. It consists of a set of 3D Gaussians, $\mathcal{\theta}_{\mathbf{O}_1}$. We are concerned with the set of 3D Gaussian means, $\mathcal{K}_{\mathbf{O}_1}=\{\boldsymbol{\mu}_i| i\in \mathcal{\theta}_{\mathbf{O}_1}\}$, where $\boldsymbol{\mu}_i=[\mu^x_i,\mu^y_i,\mu^z_i]$. 

\begin{figure}[t]
  \centering
   \includegraphics[width=\linewidth]{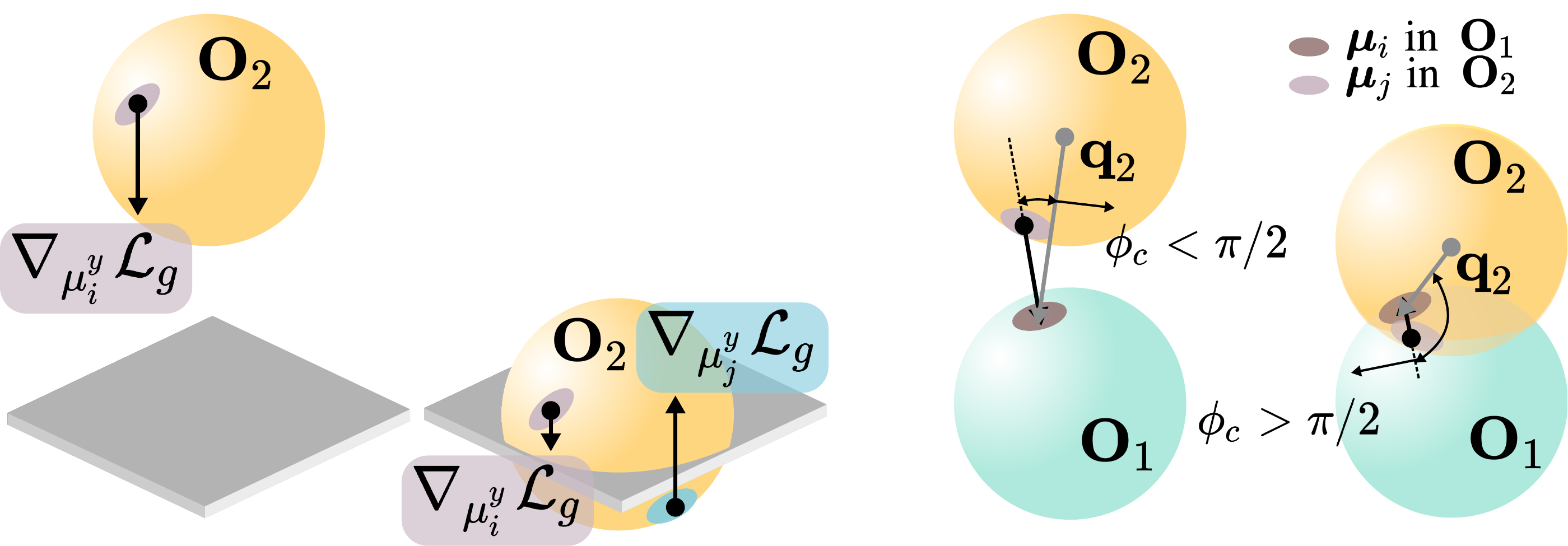}

   \begin{subfigure}[c]{0.62\linewidth}
        \centering    \caption{Gravity loss}
    \end{subfigure}
    \begin{subfigure}[c]{0.37\linewidth}
        \centering    \caption{Contact loss}
    \end{subfigure}

   \caption{\textbf{Explicit representations enable physically realistic scene composition}. Consider spherical objects, made up of several Gaussians (represented by colored ellipses). (a) The \textbf{gravity loss} provides a gradient to move the object to the virtual floor without considerably penetrating the floor. (b) The \textbf{contact loss} prevent objects from unrealistically intersecting with each other, by minimizing the angle $\theta_c$ for intersecting points.}
   \label{fig:physics_losses}
   \vspace{-0.2cm}
\end{figure}

\begin{figure*}[t]
  \centering
  \includegraphics[width=1\linewidth]{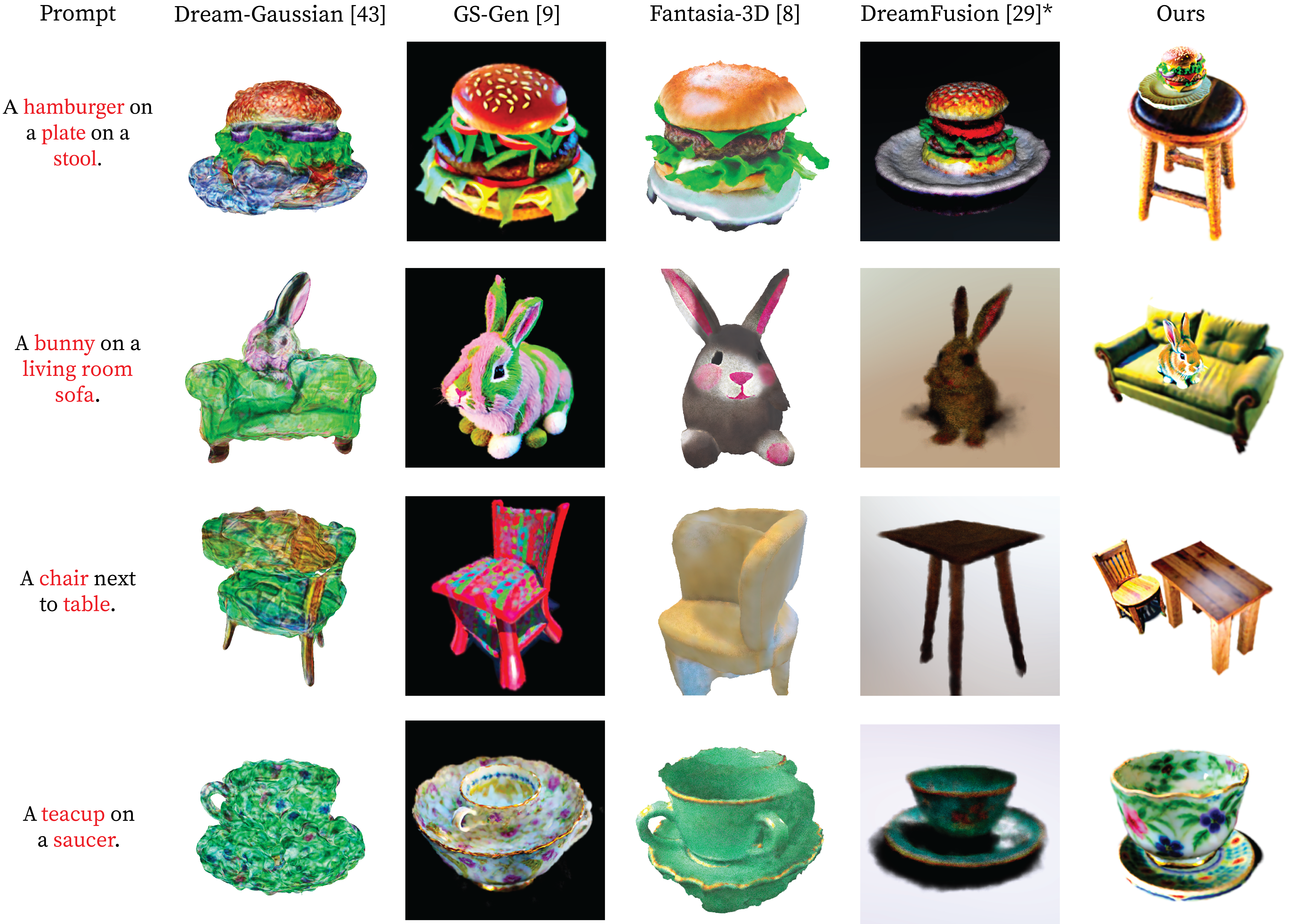}
   \caption{\textbf{Compositional generation results across diverse scenes.} For each prompt, we highlight the individual objects taken as input to our method. For DreamFusion~\cite{poole2022dreamfusion}*, we use a Stable Diffusion implementation~\cite{stable-dreamfusion} due to the original paper's proprietary model.} 
   \label{fig:big_results}
\vspace{-4mm}
\end{figure*}

\paragraph{Gravity constraint $\mathcal{L}_g$:} We define a floor for the scene as the lowermost point of anchor object $\mathbf{O}_1$. Intuitively, the gravity constraint applied on $\mathbf{O}_2$ enforces all Gaussians in $\mathbf{O}_2$ to move towards the floor without passing through it. The constraint has two distinct regimes of operation. When ${\mathbf{O}_2}$ is entirely above the floor, all Gaussians are guided to move towards the floor, via an absolute distance penalty with the floor height. When some Gaussians are below the floor, their absolute distance to the floor level is considered with a larger relative weighting, so as to pull $\mathbf{O}_1$ back above the floor. This is shown in \cref{fig:physics_losses}(a).

\paragraph{Contact constraint $\mathcal{L}_c$:} $\mathbf{O}_1$ and $\mathbf{O_2}$ are non-rigid Gaussian-represented objects. Without explicit constraints, they can intersect with each other, leading to unrealistic-looking scenes. Our key observation to enable contact constraints pertains to what we refer to as \textbf{the contact angle} $\phi^{\boldsymbol{\mu}_j}_c$ for a Gaussian $\boldsymbol{\mu}_j \in \mathcal{K}_{\mathbf{O}_2}$. For a Gaussian with mean $\boldsymbol{\mu}_j$ such that $j \in \mathbf{O}_2$, let $\boldsymbol{\mu}_i$ be the mean of the closest Gaussian in $\mathbf{O}_1$. Then, $\phi^{\boldsymbol{\mu}_j}_c$ is the angle between the vectors $\boldsymbol{\mu}_i-\mathbf{q}_2$ and $\boldsymbol{\mu}_i-\boldsymbol{\mu}_j$, where $\mathbf{q}_2$ is the center of $\mathbf{O}_2$. When $\boldsymbol{\mu}_j$ is not intersecting $\mathbf{O}_1$, $\phi^{\boldsymbol{\mu}_j}_c<\pi/2$ (\cref{fig:physics_losses}(b), left side), and when $\boldsymbol{\mu}_j$ is intersecting $\mathbf{O}_1$, $\phi^{\boldsymbol{\mu}_j}_c>\pi/2$ (\cref{fig:physics_losses}(b), right side). To avoid intersection, we enforce that the contact angle is acute for intersecting Gaussians by penalizing the negative cosine. 

The overall physics constraint is therefore given by,
\begin{equation}
    \mathcal{L}_\textrm{Phys} = \lambda_g \mathcal{L}_g + \lambda_c \mathcal{L}_c,
\end{equation}
where $\lambda_g$ and $\lambda_c$ are appropriate regularization parameters.

In addition to avoiding overlap, we provide a second contact-based constraint. Namely, if the top-view cross section areas of $\mathbf{O}_1$ and $\mathbf{O}_2$ overlap in area above a threshold, and if $\mathbf{O}_1$ and $\mathbf{O}_2$ have come into contact, $\mathbf{O}_2$ receives a small impulse toward the central axis of $\mathbf{O}_1$. This constraint compensates for limitations in our configuration intialization through SDS sampling, where objects might end up not being perfectly aligned (arising out of improper priors or 3D to 2D projective ambiguities).
Mathematical descriptions and implementation details for all the constraints may be found in the supplement.

\begin{figure*}[t]
  \centering
  \includegraphics[width=1\linewidth]{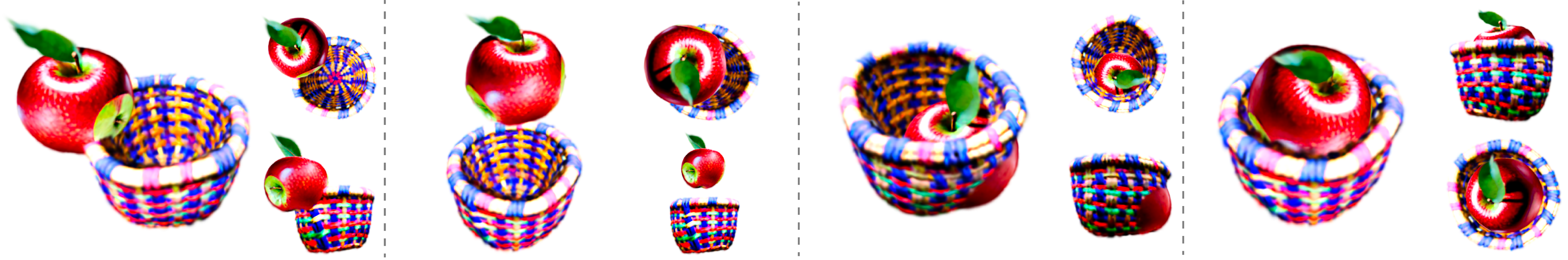}\\
  \begin{subfigure}[c]{0.24\linewidth}
    \centering    \caption{Naive SDS}
\end{subfigure}
\begin{subfigure}[c]{0.24\linewidth}
    \centering    \caption{Structured Init. (\cref{sec:sds_interaction})}
\end{subfigure}
\begin{subfigure}[c]{0.24\linewidth}
    \centering    \caption{Init. + Gravity (\cref{sec:phy_losses})}
\end{subfigure}
\begin{subfigure}[c]{0.24\linewidth}
    \centering    \caption{\textbf{Ours (Full)}}
\end{subfigure}

   \caption{\textbf{Ablation study for the composition method.} The prompt is ``a photo of a red apple in a basket''. \textbf{(a)} Naive SDS with random initialization, while \textbf{(b)} shows our initialization. \textbf{(c-d)} show the effect of the additional gravity and contact constraints.}
   \label{fig:ablation_figure}
   \vspace{-4mm}
\end{figure*}

\subsection{Text-guided object generation with Gaussian Splatting}
\label{sec:object_generation}

We sample the required objects in the composition through optimization of 3D Gaussians that can be rendered through splatting~\cite{kerbl20233d}. The primary objective is the SDS loss~\cite{poole2022dreamfusion} that we rescale according to~\cite{lin2023common} to reduce over-exposures. In addition, we use efficient geometrical initialization of the Gaussians \cite{chen2023text}, and auxillarly loss functions that distribute the Gaussians in an object in a more uniform manner. Before beginning optimization, we initialize a set of Gaussians similarly to \citet{chen2023text}, where a 3D diffusion model, Point-E \cite{nichol2022point}, generates a sparse set of points, $\mathcal{E}$, that aligns with the given prompt. The points can then be replaced by a set of isotropic Gaussians before optimization begins. However, direct optimization with the SDS objective function can produce concavities (holes) that causes blotchy object appearance. We employ K nearest neighbors (KNN) losses to distribute Gaussians more uniformely across the surface:
\begin{equation}
    \mathcal{L}_\text{KNN}(\mathcal{\theta}_1, \mathcal{\theta}_2) = \sum_{i \in \mathcal{\theta}_1}\sum_{j\in \text{KNN}(i,\mathcal{\theta}_2)} ||\mu_i-\mu_j-\text{min}(\boldsymbol{\Sigma_i})||^2,
\label{eq:knn}
\end{equation}
where $\mathcal{\theta}_1$ and $\mathcal{\theta}_2$ can be any arbitrary sets of Gaussians. To bring Gaussians to the surface we find the set of points that make up the Alpha hull of the object. A pair of points is considered part of the Alpha hull, if a line can be drawn between the pair such that a sphere of radius $1/\alpha$ contains no points in the set except for those two points on its boundary. For efficiency, we determine the set of points belonging to the Alpha hull, $\mathcal{A}$, from a subset of all Gaussians using furthest point sampling~\cite{qi2017pointnet++}.
The final objective function during the generation process is then:
\begin{equation}
    \mathcal{L}_{3D} = \mathcal{L}_{SDS} + \beta\mathcal{L}_\text{KNN}(\theta,\mathcal{A}),
\end{equation}
where $\beta$ acts as a weight for the strength of the 3D regularizer. In addition, we add the option of regularizing the locations of Gaussians to be close to the Point-E initialization, $\mathcal{L}_\text{KNN}(\mathcal{N},{\mathcal{E}})$, to preserve concavity or flatness of surfaces.

%% file: sections/4_experiments.tex
\section{Experiments}
\label{sec:discussion}
In this section, we present the results of our method to evaluate its effectiveness. We begin with \cref{ssec:zero_shot} where we show the ability of our method to generate compositional scenes while correctly predicting the pose and scale of objects with respect to each other in a semantically meaningful way. Next, in \cref{ssec:distillation}, we show that objects in a composition can be distilled for scene memory efficiency. In \cref{ssec:scene_editing} we demonstrate the flexibility our model provides in editing and recomposing scenes. We conclude with an ablation study, \cref{ssec:ablation}, that justifies our design choices. Additional results are included in the supplement. 

\paragraph{Implementation Details.} Our method uses 3D Gaussian Splatting~\cite{kerbl20233d} and  Stable Diffusion v2.1~\cite{rombach2022high}. Our method can be parallelized for $G$ GPUs such that total generation time is $30\lceil K/G \rceil + 10 P$ minutes for $K$ objects and $P$ interactions, where we use 4 Nvidia RTX3090 GPUs. The full set of hyperparameters for object generation and composition can be found in the supplement.

\subsection{Zero-shot compositional generation}
\label{ssec:zero_shot}
We compare our method's ability to generate 3D assets from language descriptions in ~\cref{fig:big_results} by comparing it to two state-of-the-art methods, Fantasia3D~\cite{chen2023fantasia3d} and DreamFusion~\cite{poole2022dreamfusion}, that use NeRF and/or mesh optimization schemes and two recently proposed methods, GS-Gen~\cite{chen2023text} and DreamGaussian~\cite{tang2023dreamgaussian}, which use Gaussian splatting. We note that the input between our method and the compared methods is different: while compared methods take in a plain text prompt, ours method takes in the text prompt decomposed into a graph, as illustrated in \cref{fig:pipeline}. That being said, baseline methods often fail at generating the expected scene. The behaviour is to  either ignore one of the items, or to fuse objects into one. For example, in row one, none of the baseline methods are successfully able to generate the stool. In the last row, we show a simpler prompt where other methods are able to generate both the saucer and teacup due to stronger diffusion guidance for the composition. Our method, generates all objects and synthesizes plausible poses and scales with respect to each other. It is important to note the detail on each object that arises out of our method, and which is notably absent from all existing baselines.


\subsection{Radiance Field Distillation.}
\label{ssec:distillation}
Due to the stochastic nature of optimizing Gaussian Radiance fields, duplication of Gaussians (splitting/cloning) is done regularly to enable expressability of the model at the cost of memory. Consequently, scenes generated with our method contain redundant Gaussians. We can distil our dense representation for each object by optimizing a new Gaussian radiance field trained on views from the original representation. We show quantitative results in \cref{tab:distillation}.

\subsection{Scene editing}
\label{ssec:scene_editing}

\begin{figure}[t]
  \centering
  \includegraphics[width=1\linewidth]{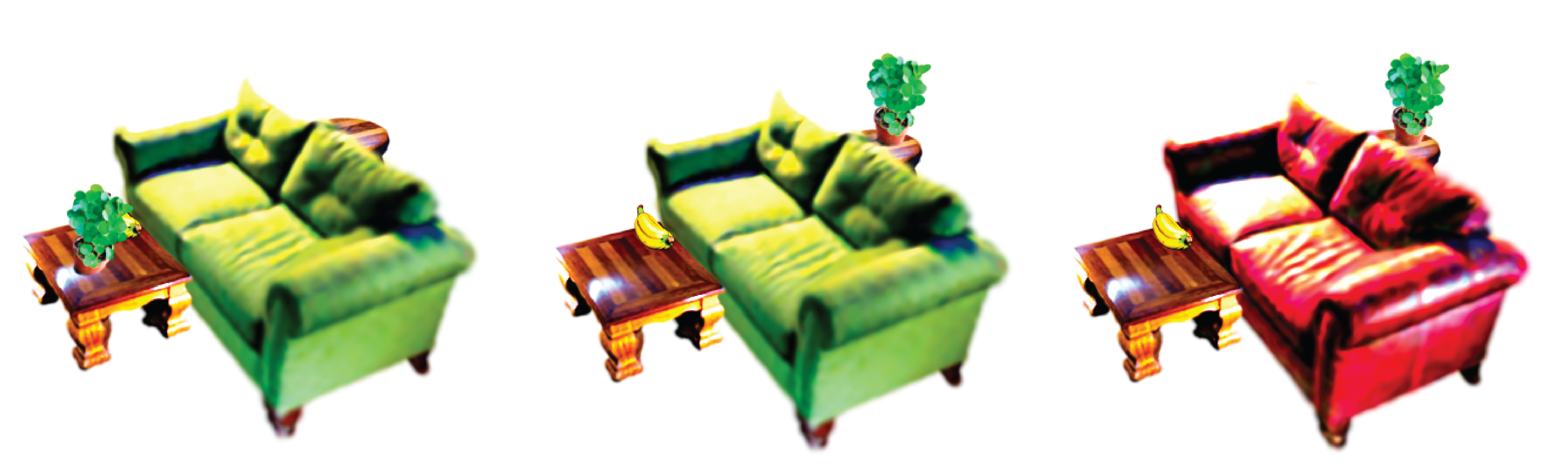}
\begin{subfigure}[c]{0.32\linewidth}
    \centering    \caption{}
\end{subfigure}
\begin{subfigure}[c]{0.32\linewidth}
    \centering    \caption{}
\end{subfigure}
\begin{subfigure}[c]{0.32\linewidth}
    \centering    \caption{}
\end{subfigure}
   \caption{\textbf{Our compositional, explicit representation allows for post-generation scene editing with high fidelity.} We use the scene from \cref{fig:teaser} to \textbf{(a):} delete the lamp from the stool, \textbf{(b) compositional editing:} move the plant from the table to the stool. \textbf{(c) object editing:} change the couch to ``a red leather couch".}
   \label{fig:scene_editing}
\vspace{-5mm}
\end{figure}

Our method offers further control over composition through scene editing. \cref{fig:scene_editing} shows several examples where our model is able to delete an object from a scene, rearrange objects in the scene and edit an individual object in the scene. Object editing is simple since each object has its own distinct parameters which we can start optimizing from for a new prompt. Recomposition requires rerunning interactions in the compositional text graph that can potentially be affected by the replacement. Thus, a user is free to recompose a full scene to create new compositions or change just one of the interactions between objects.  

\subsection{Ablation studies}
\label{ssec:ablation}

To evaluate the efficacy of our compositional method, we show the utility of each component evaluated on the composition of ``a red apple in a basket", illustrated in \cref{fig:ablation_figure}. The right-most column shows the full model that includes the initialization strategy discussed in \cref{sec:comp_generation} and the physics constraints: gravity and contact. By using only SDS for optimization with a random initialization, interactions between objects appear far from the prompt due to the poor loss landscape mentioned in \cref{sec:sds_interaction}. Our initialization, on the other hand, has higher agreement with the given prompt. However, without the gravity constraint, we find that objects may often appear to be suspended in mid-air since the diffusion model may find this to be a plausible composition as illustrated in \cref{fig:physics_accuracy}. The contact loss is important for preserving each object's individuality and to eliminate cases where objects are within each other.  All components are vital to create physically realistic scenes and overcome biases in text-to-image diffusion models. 

\begin{table}[b]
    \centering
    \footnotesize
    \begin{tabular}{c>{\centering\arraybackslash}p{1.7cm}>{\centering\arraybackslash}p{1.7cm}c}
         \toprule
         Object & Original (\#Gaussians) & Distilled (\#Gaussians) & PSNR (dB)\\
         \midrule
        Table & 41,928 & 6,872 & 44.16 \\
        Teacup & 48,370 & 15,856 & 39.22 \\
        Hamburger & 51,937 & 21,955 & 36.57\\
        \bottomrule
    \end{tabular}
    \caption{\textbf{Distillation of Gaussian radiance field.} For several objects from the \cref{fig:big_results} compositions, we report the number of Gaussians and PSNR between the original and distilled representations.} 
    \label{tab:distillation}
\end{table}

\begin{figure}[t]
  \centering
  \includegraphics[width=0.7\linewidth]{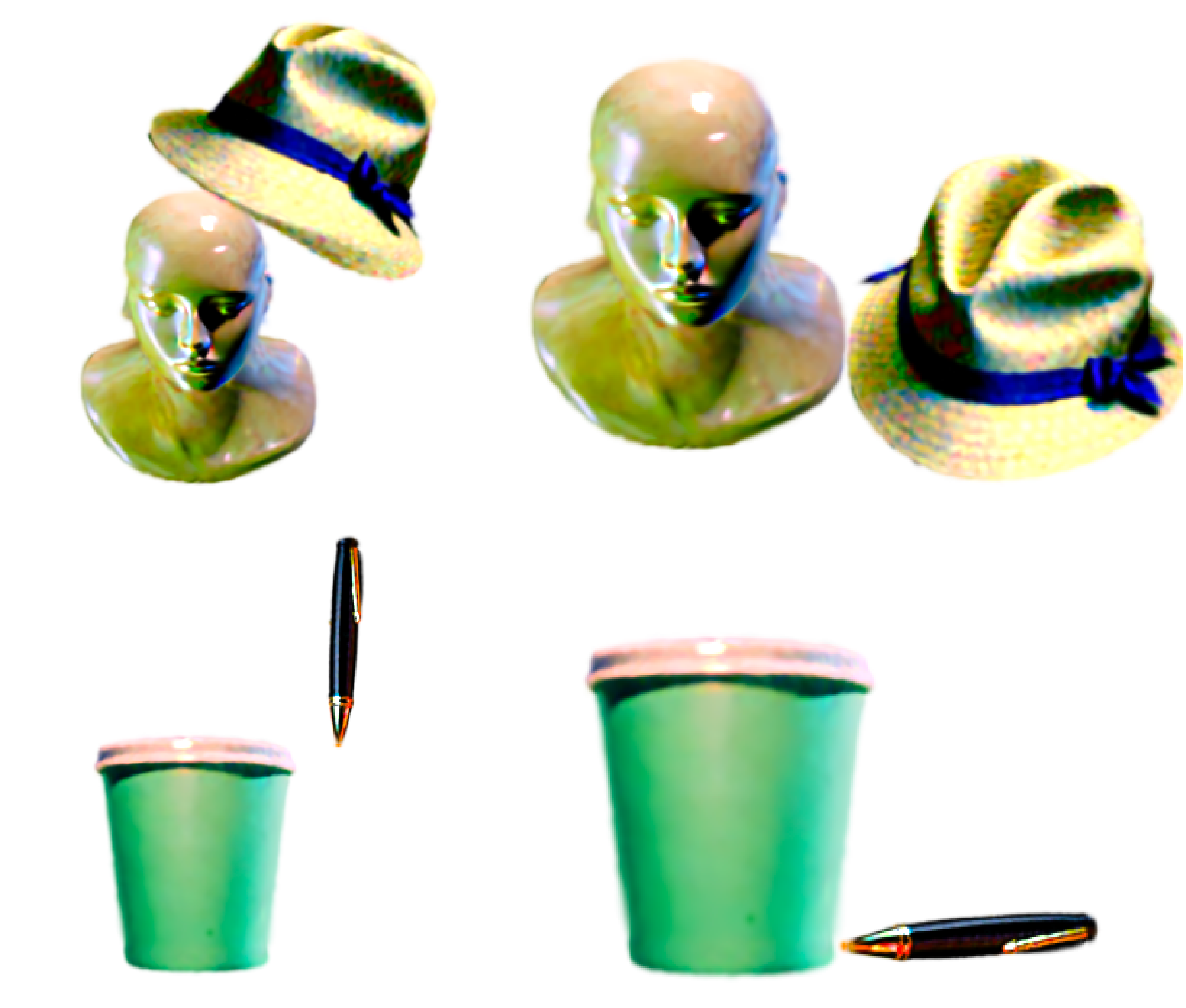}
\begin{subfigure}[c]{0.1\linewidth}
    \centering  
\end{subfigure}

\begin{subfigure}[c]{0.32\linewidth}
    \centering    \caption{Failure conditions}
\end{subfigure}
\begin{subfigure}[c]{0.36\linewidth}
    \centering    \caption{Final result}
\end{subfigure}

\begin{subfigure}[c]{0.2\linewidth}
    \centering  
\end{subfigure}

   \caption{\textbf{Our physical constraints and initialization can fail under certain conditions.} The top and bottom rows show runs for ``a hat on a mannequin" and ``a black pen in a office cup".}
   \label{fig:limitations}
\vspace{-5mm}
\end{figure}

%% file: sections/5_limitations.tex
\section{Limitations and Open Problems}
\label{sec:limitations}

 3D scene generation from text is a difficult problem. While we propose a new framework for compositional generation, several limitations still exist. Our method assumes rigid-body interactions during composition, hence we do not support interactions where objects need to go into each other or require deformations. Here, we discuss two specific examples. We show in \cref{fig:limitations}, a failure of an attempt at ``a hat on a mannequin" where the hat slips off due to not perfectly fitting the head, and the lack of frictional forces in our current physical constraints. Additionally, our method is sensitive to initialization. For example, as shown in the ``pen in a cup" scene, the pen had a poor initialization that led it to not being in the cup. In this case, the failure arises as a result of our scheme to address the translation anomaly in the SDS loss~\cref{sec:sds_interaction}. Our configuration estimation is also susceptible to SDS guidance: any anomalies in the diffusion model guidance will therefore reflect as failure cases. Our method also inherits common object generation problems such as the Janus problem~\cite{armandpour2023re} and geometry/texture ambiguities that may potentially confuse composition generation. Our method does not solve these issues, but several recent works can alleviate them~\cite{armandpour2023re, seo2023let, xu2023dream3d}. 

%% file: sections/6_conclusion.tex
\section{Conclusion}
\label{sec:Conclusion}
We propose \paperAcronym, a method for generating compositional 3D scenes using only text. We aim to fill the gap where prior methods often fail to generate coherent multi-object scenes. By utilizing explicit radiance fields, our method can create physically correct scenes, while maintaining object individuality and fidelity. Generated compositions can be edited and rearranged in under 15 minutes through text prompts alone, thus giving users flexibility and freedom to create more diverse scenes in less time. In future research, we aim to enable more elaborate interactions between objects and support large scale compositions with orders of magnitude higher number of objects.

\paragraph{Ethics Statement:} Generative models have the potential to create harmful content. We condemn any such use for malicious purposes.

%% file: sections/A_suppl.tex
\clearpage
\setcounter{section}{0}
\setcounter{figure}{0}
\maketitlesupplementary

\renewcommand\thesection{\Alph{section}} 
\renewcommand\thefigure{\Alph{figure}} 
\renewcommand\thetable{\Alph{table}} 

\section{Supplemental Contents}
\label{sec:contents}
This supplement is organized as follows:
\begin{itemize}
    \item Score Distillation Sampling for Composition Guidance
    \item Incorporating Physical Losses for Composition
    \item Object Generation Implementation Details
    \item User Input Details
    \item Additional Results
\end{itemize}

\section{Score Distillation Sampling for Composition Guidance}
\label{sec:sds_analysis}

\subsection{The scale anomaly}
Secition 3.1.1 covers an initial description of the scale anomaly for configuration using SDS. Here, we will cover it in additional detail and explore additional aspects of the anomaly.

For the purpose of continuity, we will begin by redefining the anomaly. We refer to this anomaly as the monotonic decrease in the configurational function $\mathcal{F}$ with decreasing scale $s_{2,1}$. (\textbf{Note:} In the main paper, we refer to the function $\mathcal{F}$ as the configuration likelihood function, however the desired objective for this function is minimization rather than maximization (which is the case for likelihood functions usually).)

\begin{figure}[h]
    \centering
    \includegraphics[width=\linewidth]{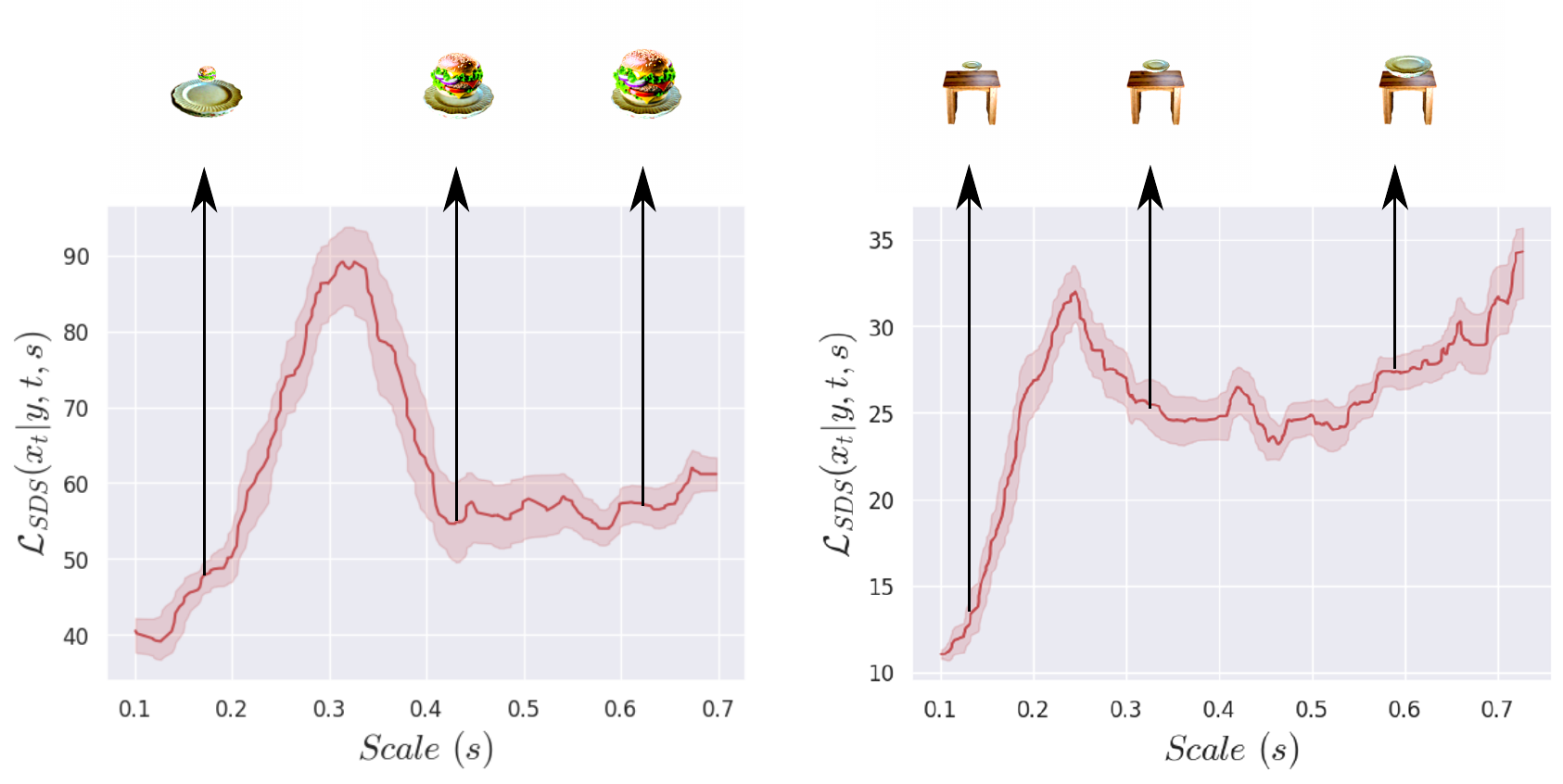}
    \caption{\textbf{The scale anomaly exists across different object combinations.} For a fixed camera radius, the threshold below which the anomaly occurs, and nature of the anomaly are broadly consistent.}
    \label{fig:scale_object}
\end{figure}

\paragraph{Effect of object variation:} Figure~\ref{fig:scale_object} shows the anomaly for two different object configurations. We find the scale anomaly to be broadly consistent across objects. We also note that the threshold below which the scale anomaly occurs is broadly independent of the object, and depends on the relative size of the object in the rendered image. Since our object generation step generates objects of broadly the same size, the threshold remains constant regardless of the object.

\begin{figure}[h]
    \centering
    \includegraphics[width=\linewidth]{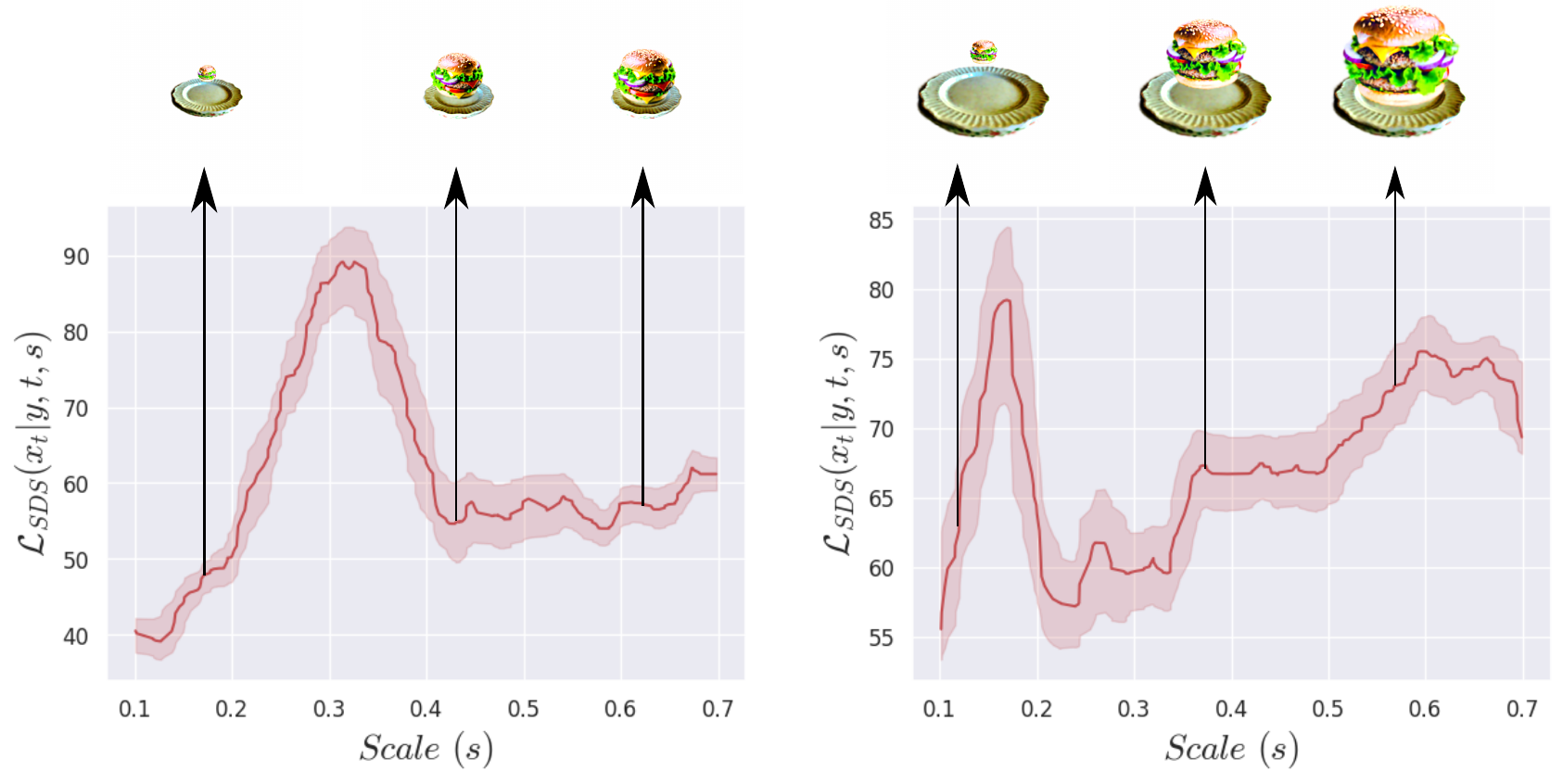}
    \caption{\textbf{Effect of camera radius on the scale anomaly.} We find that the threshold for the scale anomaly moves to a lower scale value as the camera radius becomes smaller.}
    \label{fig:scale_radius}
\end{figure}

\paragraph{Effect of camera radius:} Figure~\ref{fig:scale_radius} shows the anomaly for two different camera radii. We find that the scale anomaly is affected by the radius. The smaller the camera radius from the center of the object, the lower is the scale below which the anomaly is seen. This observation is consistent with the hypothesis that the anomaly occurs when the relative object size is small in the rendered image.

\paragraph{Practical limitations for composition:} The scale anomaly enforces limitations on SDS-based configuration estimation. Specifically, (a) if the camera radius is large enough, below a certain scale the $\mathcal{F}$ function will not provide accurate guidance (as a result of the scale anomaly), and (b) if the camera radius is small enough, above a certain scale the $\mathcal{F}$ function will not provide accurate guidance (since $\mathbf{O}_2$ will not be entirely visible in the rendered image).


\subsection{The translation anomaly}
\label{sec:trans_anomaly}
Section 3.1.1 covers an initial description of the translation anomaly for configuration using SDS. Here, we will cover it in additional detail and explore additional aspects of the anomaly.

The translation anomaly is the reduction in the configurational function $\mathcal{F}$ for regions with object occlusion. That is, the value of $\mathcal{F}$ is anomalously lower if certain parts of the object are not visible to the camera.

\begin{figure}[h]
    \centering
    \includegraphics[width=\linewidth]{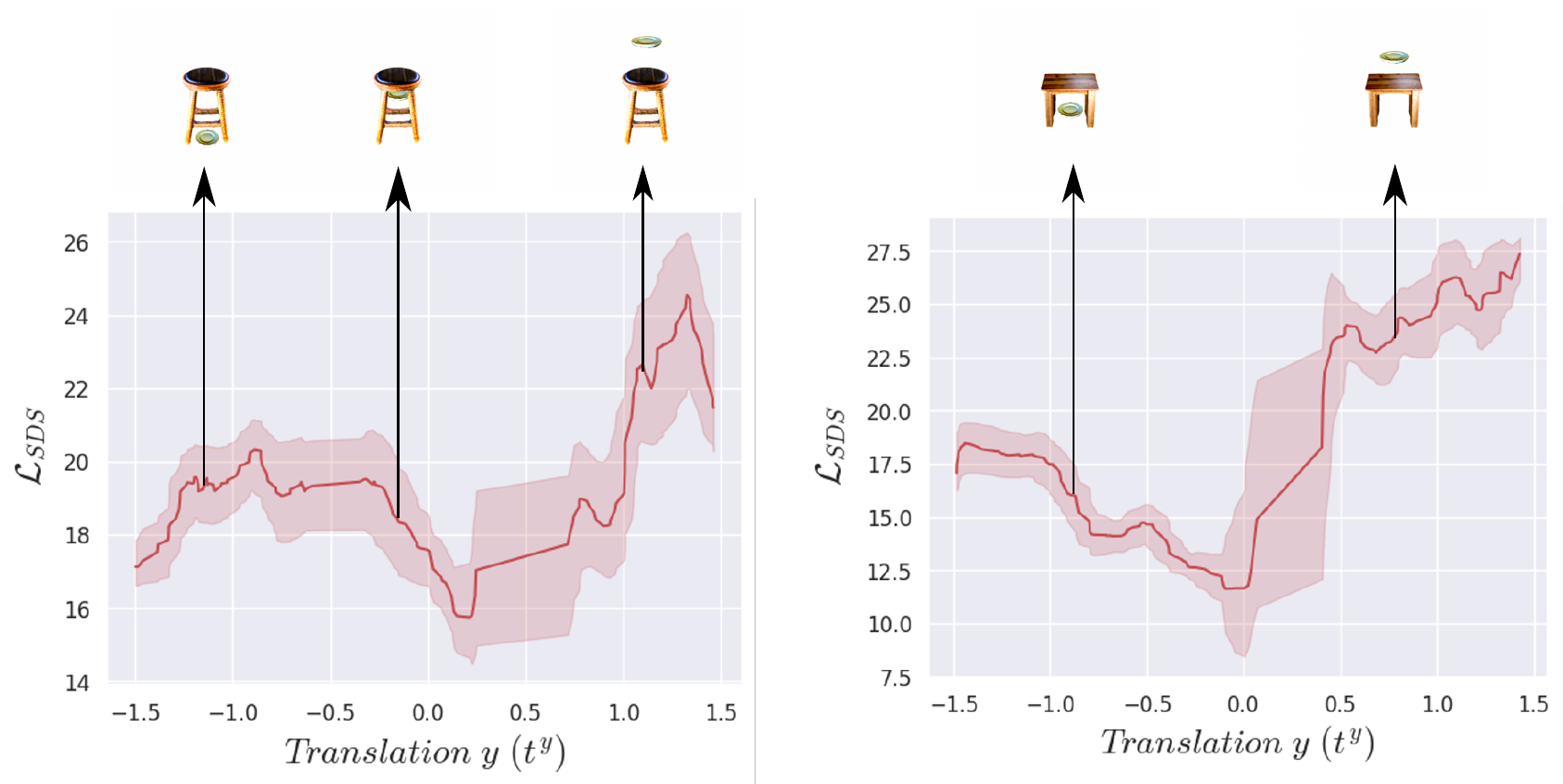}
    \caption{\textbf{The translation anomaly for composition guidance.} The SDS loss observes lower values for regions with occlusion, such as below a table or a stool. This can lead to false optima in these occluded regions.}
    \label{fig:trans_object}
\end{figure}

\paragraph{Effect of occlusion:} Figure~\ref{fig:trans_object} shows the translation anomaly for three different object configurations. We find that the translation anomaly qualitatively exists across various object configurations. However, the nature of the anomaly varies across configurations in terms of the extent of the anomaly.

\paragraph{Practical limitations for composition:} The following limitations to composition estimation are enforced by the translation anomaly: (a) occluded scenes (eg. a pen in a cup, a corgi under a table etc.) do not have accurate guidance from the $\mathcal{F}$ function, (b) the scale of the object affects the translation anomaly. Therefore, the $\mathcal{F}$ function would favor smaller, occluded objects as a result of the optimization. 

\subsection{Camera positioning}
\label{sec:camera_positioning}
\begin{figure}[h]
  \centering
   \includegraphics[width=0.49\linewidth]{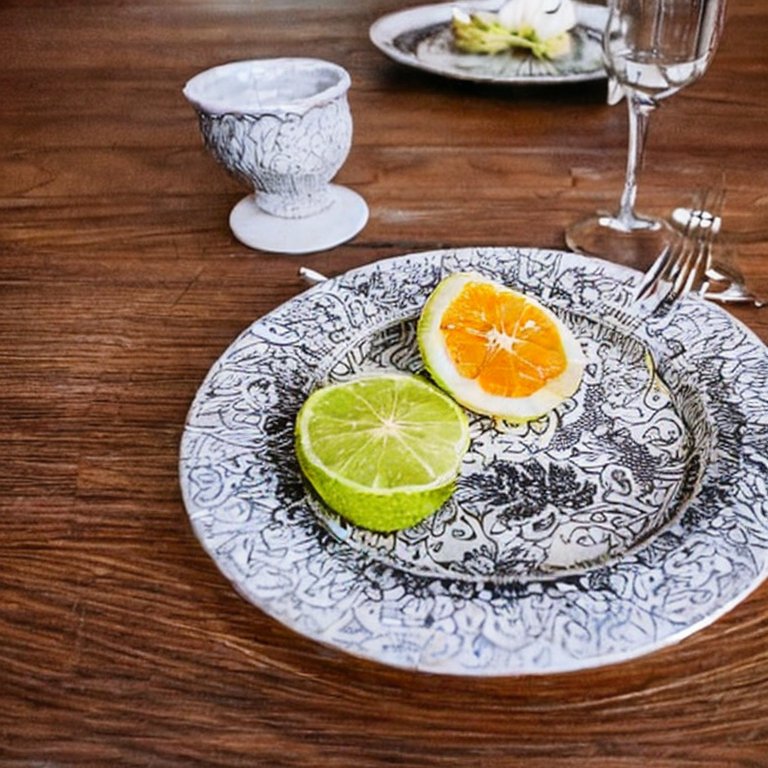}
   \includegraphics[width=0.49\linewidth]{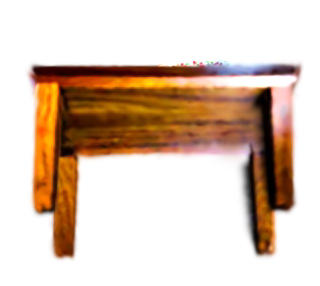}
    \caption{\textbf{Our base diffusion model, Stable Diffusion v2.1~\cite{rombach2022high}, prefers positive camera elevations (looking down from up).} (Left) For the prompt ``a plate on a table", the generative prior considerably prefers top-down views. Therefore, for our composition, we do not use camera views with negative elevation (looking up from down) since these views do not have significant support in the domain of the generative prior (as shown on the right, for the scene ``a plate on a table").}
   \label{fig:camera_view}
   
\end{figure}

Another aspect that can affect the behavior of SDS-based configuration optimization is the camera positioning and view perspectives used. To understand this, it useful to understand canonical views that a diffusion-based image generation model is inclined towards generating, from the perspective of combinations of multiple objects. Figure~\ref{fig:camera_view} shows an image generated by the Stable Diffusion v2.1 model. Note how camera perspective has a positive elevation (looking down from up) in most compositional scenes. This is consistent with our observations that positively elevated camera angles lead to more stable composition.

\paragraph{Negative camera elevations:} From the perspective of loss function design, it is intuitively beneficial to have symmetric camera positions. Therefore, along with positive camera elevations, we would ideally also prefer to have negative camera elevations. This might also help with addressing the translation anomaly. However, there is a problem. Figure~\ref{fig:camera_view} shows an example view from a negative elevation: such views are not distinctive of the scene prior in the diffusion model, and therefore do not contain signal for compositional guidance. Therefore, we decide to limit our camera views to contain only positive camera elevation angles. Specifically, for our structured initialization, we use 8 fixed images, split across two elevations (30 and 60 degrees) and four azimuths (0, 90, 180 and 270 degrees).

\subsection{Our structured initialization, in detail}
\label{sec:detailed_init}
As mentioned in the main paper, to address the aforementioned anomalies, simple Monte Carlo sampling is insufficient to identify an initial compositional configuration that is faithful to the given text prompt $\mathbf{y}$. At a high level we perform an alternating sampling of translation and scale, while keeping rotation fixed (the rotation gets optimized subsequently through the physics and SDS loss updates post initialization). That is,
\begin{enumerate}
    \item First, joint initialization of scale and translation
    \item Keeping scale fixed, choose a suitable translation
    \item Keep translation fixed, choose a suitable scale
    \item Repeat the above two steps $L=3$ times, while keeping the rotation fixed
\end{enumerate}
\cref{alg:init} describes the structured initialization. We now explore both scale and translation initialization in detail.

\begin{algorithm}[t]
\caption{Our structure initialization}\label{alg:init}
\begin{algorithmic}
\Require $\mathbf{O}_1,\mathbf{O}_2,\mathbf{y}_{1,2}$, M.C. sampler $\textrm{\textbf{MonteCaro}}\{\cdot\}$ \Comment{Objects, interaction text description, sampling operators}
\Ensure $\mathbf{R}^*_{2,1},\mathbf{t}^*_{2,1},s^*_{2,1}$ \Comment{Interaction parameters}
\State $N \gets 3$ \Comment{\# of alternating optimization steps}
\State $n \gets 1$ \Comment{Current step}
\State $\mathbf{R}^*_{2,1}=[1,0,0,0]$ \Comment{Fix rotation to identity quaternion}

\State $\mathbf{t}^*_{2,1},s^*_{2,1} = \underset{\mathbf{t}_{2,1},s_{2,1}}{\textrm{\textbf{MonteCarlo}}}\{\mathcal{F}_\text{Trans}(\mathbf{R}^*_{2,1},\mathbf{t}_{2,1},s_{2,1})\}$ \Comment{Joint}

\While{$n \leq N$}
\State $\mathbf{t}^*_{2,1} = \underset{\mathbf{t}_{2,1}}{\textrm{\textbf{MonteCarlo}}}\{\mathcal{F}_\text{Trans}(\mathbf{R}^*_{2,1},\mathbf{t}_{2,1},s^*_{2,1})\}$ \Comment{Transl.}

\State $s^*_{2,1} = \underset{s_{2,1}}{\textrm{\textbf{MonteCarlo}}}\{\mathcal{F}(\mathbf{R}^*_{2,1},\mathbf{t}^*_{2,1},s_{2,1})\}$ 

\Comment{Scale}

\State $n \gets n+1$
\EndWhile
\end{algorithmic}
\end{algorithm}

\subsubsection{Scale initialization} The scale anomaly is broadly consistent across object-pairs and depends on the camera radius for rendering (\cref{fig:scale_object,fig:scale_radius}). We find that the most stable approach therefore is to limit the candidate range of scales to exclude the anomalous region, for a given camera radius. In practice, we begin scale sampling at a camera radius of 4.5 units, by identify 50 random scales. The anchor object $\mathbf{O}_1$ has a fixed scale of 0.8, while the set of candidate scales for the section object $\mathbf{O}_2$ are chosen such that $s_{2,1}\in [0.3,0.7]$. The desired scale is chosen as the average of the 5 scales with least values of the CLF $\mathcal{F}$. If the chosen scale is found to be within a threshold of 0.05 of the range lower limit (that is, if $s_{2,1}<0.35$), we reduce the camera radius to 2.5 units and change the candidate scale range to $[0.2,0.6]$. The lower range limits are chosen and fixed based on our scale analysis across several objects, and we find this approach to be more stable that one involving and adaptive estimate of the range lower limit (as a result of the noisy nature of the CLF $\mathcal{F}$). Therefore, our overall effective range of candidate scales is $[0.2,0.7]$ with the anchor object $\mathbf{O}_1$ having a scale of 0.8. We find this range to be sufficient to accommodate a variety of composition requirements as evidenced through our results.

\subsubsection{Translation initialization} 
\paragraph{Visibility function:} For effective Monte Carlo search across translation, we need to account for the translation anomaly. To achieve this, we design a representative function for object visibility. Specifically, we use the viewspace gradients available within the 3D Gaussian Splatting framework~\cite{kerbl20233d}. Given an objective function, the viewspace gradients essentially supply the dependence of the objective on individual Gaussians. By defining said objective to be the rendered image (averaged across pixels and noised, for conditioning), we can get the contributions of each Gaussian towards the image. Averaging for all Gaussians in $\mathbf{O}_2$ is then our measure of object visibility $v(\cdot)$. 
There remains a problem with this visibility function. Since it is a pure measure of object contribution to an image, apparent visibility will reduce as the object is farther away from the cameras. While this can be solved with symmetric cameras (in terms of elevation), we are unable to do that due to the nature of the diffusion guidance~\cref{sec:camera_positioning}. We find that defining the visibility function with an exponent $\gamma<1$ reduces this effect favorably, in practice. Additionally, to nullify the dependence of the effective visibility function on the scale of the object (larger objects occupy more pixels in the image), we normalize with the square of the scale $s_{2,1}^2$. Therefore, our effective CLF for translation becomes,
\begin{equation}
    \mathcal{F}_\text{Trans}=\mathcal{F}\cdot/\left(\frac{v}{s^2_{2,1}}\right)^{-\gamma},
\end{equation}
where $v$ is the visibility function defined by us, in terms of he viewspace gradients, $s_{2,1}$ is the scale of $\mathbf{O}_2$, $\mathbf{F}$ is our originally defined CLF and $\gamma$ is the exponent.

\paragraph{The initialization regime:} At each step, we sample 50 points on a sphere around the object $\mathbf{O}_1$, to get candidate translations $\mathbf{t}_{2,1}$ (we choose this as $\mathbf{O}_1$ as generated with our object generation is smaller than this radius). Candidate translations are pruned to ignore samples where the two objects intersect, and when $\mathbf{O}_2$ is below the floor (lowermost point of $\mathbf{O}_1$). Post this, the best translation is chosen such that it minimizes the updated translation CLF $\mathcal{F}_\text{Trans}$.

\subsubsection{Joint initialization}
To accommodate joint initialization, we combine aspects of both scale and translation initialization. We sample 150 $\mathbf{t}_{2,1}-s_{2,1}$ combinations, being consistent with the smapling constraints of translation (pruning floor violations and intersections) and scale (limiting scale sampling to within the radius-specific range). We use the translation CLF $\mathcal{F}_\text{Trans}$ to identify the combination with highest likelihood. We do not average across the 5 best samples, as we do for scale initialization.

\subsection{Success Rate of Composition}
We find different object configurations to have different rates of success with regards to composition. We summarize our observations below.

\subsubsection{Dependence on Diffusion Guidance}
We find strong diffusion guidance to be the most important factor in determining the success of our composition step. We are therefore limited to object configurations and pairs for which the base diffusion model has a strong prior, and hence strong guidance.

\subsubsection{Variance across object instance}

\begin{figure}[h]
  \centering
   \includegraphics[width=0.49\linewidth]{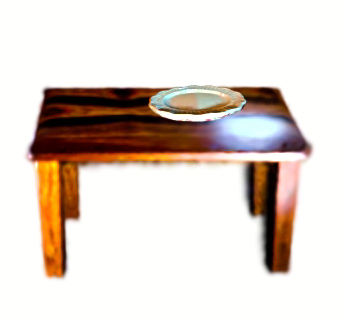}
   \includegraphics[width=0.49\linewidth]{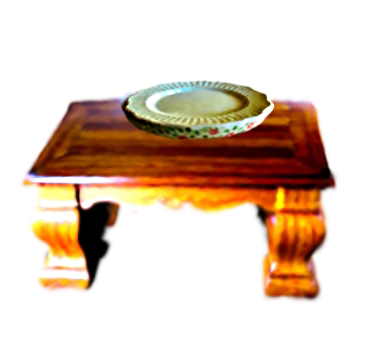}
    \caption{\textbf{The specific object instances can affect the composition parameters.} The apparently larger appearance of the table on the left (as a result of slender legs, etc.) leads to a smaller overall plate relative to it. The smaller appearance of the table on the right leads to a relatively larger plate.}
   \label{fig:obj_variance}
   
\end{figure}

Given a fixed prompt, score-based composition guidance can show considerable variation as a function of the individual objects. Here, we show two examples of composition for the prompt ``a photo of a plate on a table". Specifically, we apply or composition technique across two different pairs of plates and tables. Across three trials, we consistently observe that our guidance infers a considerably larger size for one of the plates on the table, as opposed to the other. In other words, depending on the specific instance of a plate and a table, the SDS loss has a variable interpretation of their relative scales that it enforces. Figure~\ref{fig:obj_variance} highlights this observation.

\subsubsection{Close-up Versus Long-shot Compositons}
\begin{figure}[h]
  \centering
   \includegraphics[width=0.49\linewidth]{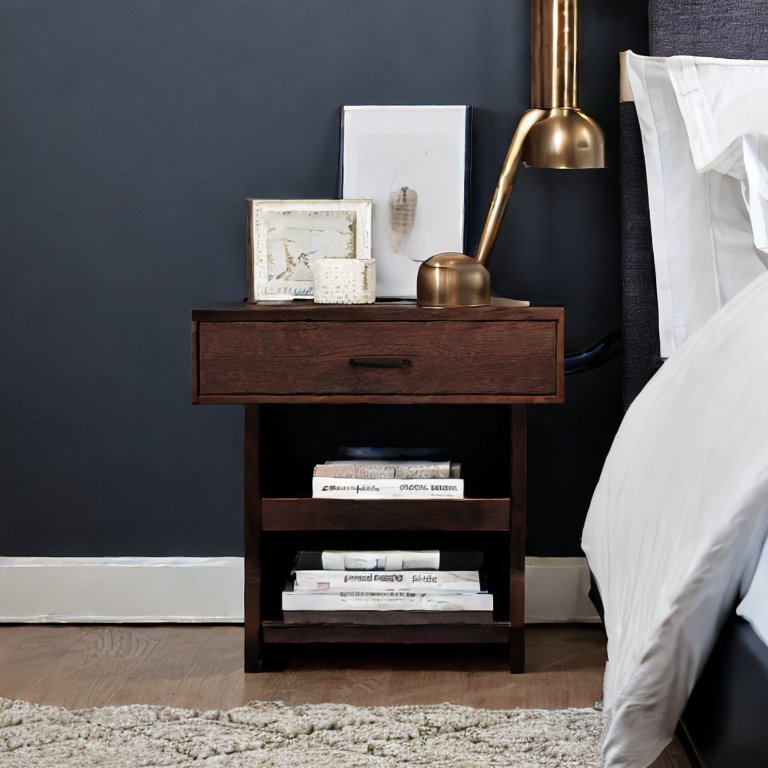}
   \includegraphics[width=0.49\linewidth]{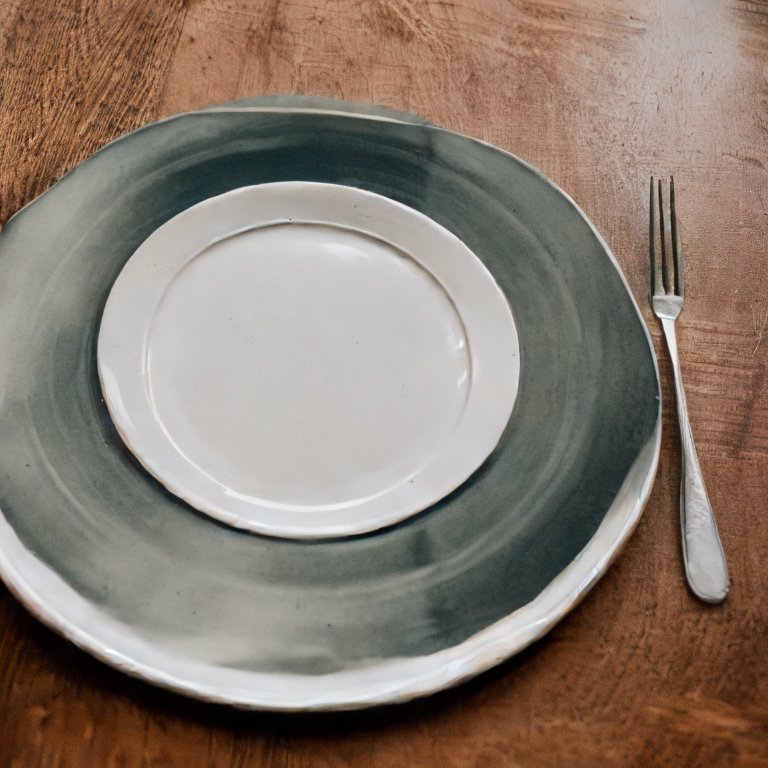}
    \caption{\textbf{Our base diffusion model, Stable Diffusion v2.1~\cite{rombach2022high}, prefers object-centric, close-up views of scenes, which may hinder guidance for larger-scale scene composition.} (Left) ``Wide shot, a night stand next to a bed". (Right) ``wide shot, a plate on a table". In both cases, the second object (bed, table) does not have good visibility.}
   \label{fig:close_up}
   
\end{figure}

We find that the Stable Diffusion v2.1 guidance favors close-up camera views centered on the primary object, even for wider-scale scenes. Figure~\ref{fig:close_up} shows two room-scale scenes, for which the image remains relatively closer-up. In terms of composition, we find this to manifest in the form of more reliable guidance for close-up scenes (``a hamburger on a plate") than for wide-angle scenes (``a nightstand next to a bed").

\subsubsection{Dependence on object-level Janus Effect}

\begin{figure}[h]
  \centering
   \includegraphics[width=0.49\linewidth]{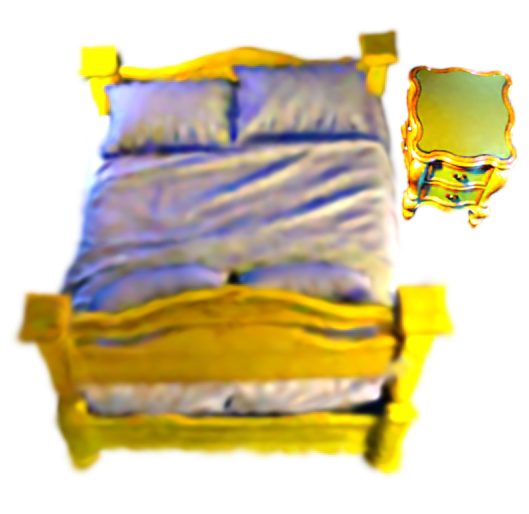}
   \includegraphics[width=0.49\linewidth]{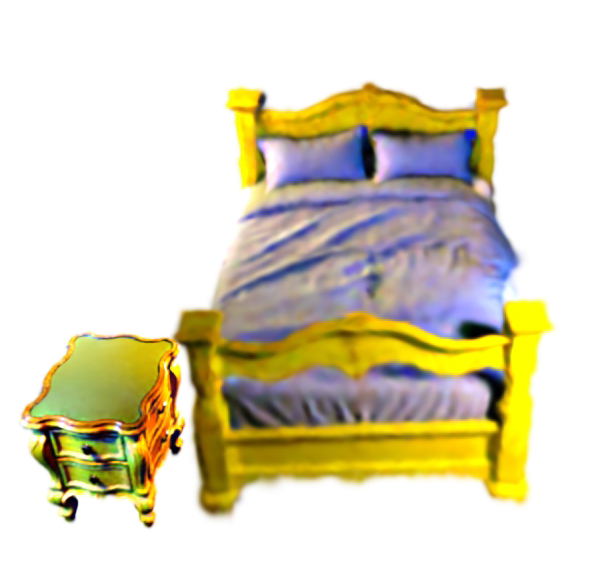}
    \caption{\textbf{For the above scene ``a nightstand next to a bed", object-level Janus effects can lead to multiple possible solutions for composition.} (Left) A viable composition for the scene, given the prompt. (Right) However, as a result of the Janus effect, rotating the azimuthal angle by 180$^{\circ}$ shows the possibility for a different viable configuration (with the nightstand further up along the bed from this perspective).}
   \label{fig:janus_effects}
   
\end{figure}

Object-level Janus effects can lead to multiple likely configurations. We show an example of this in Figure~\ref{fig:janus_effects} shows the scene ``a nightstand next to a bed". Janus effects on the bed can lead to multiple likely configurations, out of which one is arrived at through our composition.

\subsubsection{Objects with Occlusions}
As discussed in Section~\ref{sec:trans_anomaly}, quirks in the SDS guidance lead to false optima at locations with occlusions (like ``a pen in a cup"). Therefore, we are unable to support such configurations with occlusions through our guidance.

\section{Physical Losses for Composition}
\label{sec:phy_losses_supp}
We now describe the physics-based optimization step to enable realistic composition (\cref{sec:phy_losses}, main paper). As described there, post the structured initialization, we aim to optimize the following loss function:
\begin{equation}
    (\mathbf{R}^*_{2,1},\mathbf{t}^*_{2,1},s^*_{2,1})=\underbrace{\underset{\mathbf{R}_{2,1},\mathbf{t}_{2,1},s_{2,1}}{\arg \min} \mathcal{F}+\mathcal{L}_\text{Phys}\big\vert_{\textrm{init. at }\mathbf{P}'_{2,1}}}_{\text{Phys. finetune}}.
\end{equation}
That is, after a structured initialization, we optimize the rotation, translation and scale of $\mathbf{O}_2$ with respect to $\mathbf{O}_1$ using the CLF regularized by physics losses.

Consider object $\mathbf{O}_1$. It consists of a set of 3D Gaussians, $\theta_{\mathbf{O}_1}$. We are concerned with the set of 3D Gaussian means, $\mathcal{K}_{\mathbf{O}_1}=\{\boldsymbol{\mu}_i| i\in \theta_{\mathbf{O}_1}\}$, where $\boldsymbol{\mu}_i=[\mu^x_i,\mu^y_i,\mu^z_i]$. Based on this, the geometric center is defined as $\mathbf{q}_1 = \frac{1}{|\theta_{\mathbf{O}_1}|}\underset{\boldsymbol{\mu}_i\in \mathcal{K}_{\mathbf{O}_1}}{\sum \boldsymbol{\mu}_i}$, where $|\cdot|$ is the cardinality operator. Similarly, we can define $\mathcal{K}_{\mathbf{O}_2}$ and associated parameters. The `up' direction is along the $y$-axis.

\begin{figure}[t]
  \centering
   \includegraphics[width=\linewidth]{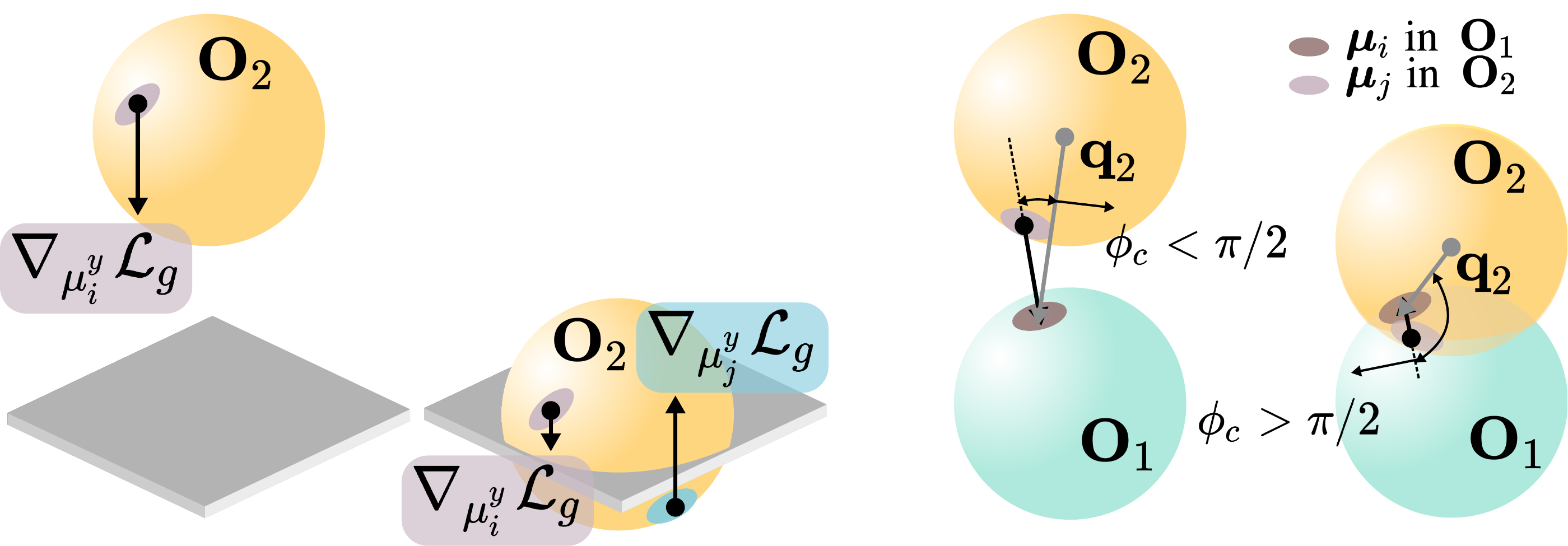}

   \begin{subfigure}[c]{0.62\linewidth}
        \centering    \caption{Gravity loss}
    \end{subfigure}
    \begin{subfigure}[c]{0.37\linewidth}
        \centering    \caption{Contact loss}
    \end{subfigure}

   \caption{\textbf{Explicit representations enable physically realistic scene composition}. Consider spherical objects, made up of several Gaussians (represented by colored ellipses). (a) The \textbf{gravity loss} provides a gradient to move the object to the virtual floor without considerably penetrating the floor. (b) The \textbf{contact loss} prevent objects from unrealistically intersecting with each other, by minimizing the angle $\theta_c$ for intersecting points. Repeated from main paper for clarity.}
   \label{fig:physics_losses_supp}
   \vspace{-0.2cm}
\end{figure}

\paragraph{Gravity loss $\mathcal{L}_g$:} We define a floor for the scene as the lowermost point of $\mathbf{O}_1$. Specifically,
\begin{equation}
    c^{\text{floor}}=\min \mu^y_i \, \forall  \,\mathbf{\mu}_i \in \mathcal{K}_{\mathbf{O}_1}.
\end{equation} 
In practice, we choose the floor as the median of the y-coordinates of the lowest 0.1\% of Gaussians in $\mathbf{O}_1$. Intuitively, the gravity loss applied on $\mathbf{O}_2$ must enforce all Gaussians in $\mathbf{O}_2$ to move towards the floor without passing through it. That is, the loss has two distinct regimes of operation. When ${\mathbf{O}_2}$ is entirely above the floor,
\begin{equation}
    \mathcal{L}_\textrm{g} =
      \frac{1}{|\mathcal{K}_{\mathbf{O}_2}|}\underset{\boldsymbol{\mu}_i \in \mathcal{K}_{\mathbf{O}_2}}{\sum} \left(\mu^y_i-c^{\textrm{floor}}\right) \text{    if $\mu^y_i>c^{\textrm{floor}}\,\forall \, \boldsymbol{\mu}_i \in \mathcal{K}_{\mathbf{O}_2}$}.
\end{equation}
This is shown in \cref{fig:physics_losses_supp}(a), left side. However, when some Gaussians are below the floor, it should provide a larger relative gradient to pull those Gaussians back above the floor. That is, in such a setting, the loss is given by,
\begin{multline}
    \mathcal{L}_\textrm{g} =
      \frac{1}{K_{\textrm{comb}}}\left[\frac{1}{|{\mathcal{K}_{\mathbf{O}_2}}^+|}\underset{\boldsymbol{\mu}_i \in {\mathcal{K}_{\mathbf{O}_2}}^+}{\sum} \left(\mu^y_i-c^{\textrm{floor}}\right)\right]\\
      +\frac{1}{|{\mathcal{K}_{\mathbf{O}_2}}^-|}\underset{\boldsymbol{\mu}_i \in {\mathcal{K}_{\mathbf{O}_2}}^-}{\sum} \left(c^{\textrm{floor}}-\mu^y_i\right),
      \label{eq:grav_loss_pt1}
\end{multline}
where $K_\textrm{comb}>>1$ is a combination factor between the two terms, and ${\mathcal{K}_{\mathbf{O}_2}}^+$ and ${\mathcal{K}_{\mathbf{O}_2}}^-$ are defined such that,
\begin{equation}
\begin{split}
    {\mathcal{K}_{\mathbf{O}_2}}^+ &= \{\boldsymbol{\mu}_i \,\,|\,\, \boldsymbol{\mu}_i \in \mathcal{K}_{\mathbf{O}_2},\,\mu^y_i>c^{\textrm{floor}}\}\\
    {\mathcal{K}_{\mathbf{O}_2}}^- &= \{\boldsymbol{\mu}_i \,\,|\,\, \boldsymbol{\mu}_i \in \mathcal{K}_{\mathbf{O}_2},\,\mu^y_i<c^{\textrm{floor}}\}.
\end{split}
\label{eq:grav_loss_pt2}
\end{equation}
This is shown in \cref{fig:physics_losses_supp}(a), right side. In practice, we use $K_{\textrm{comb}}=2000$. 

\paragraph{Contact loss $\mathcal{L}_c$:} $\mathbf{O}_1$ and $\mathbf{O_2}$ are non-rigid Gaussian-represented objects. Without explicit constraints, they can intersect with each other, leading to unrealistic-looking scenes. The contact loss addresses this. Enforcing such a loss is nuanced. Gaussian-represented objects do not have a defined surface, along which contact occurs. Our key observation pertains to what we refer to as \textbf{the contact angle} $\theta^{\boldsymbol{\mu}_j}_c$ for a Gaussian $\boldsymbol{\mu}_j \in \mathcal{K}_{\mathbf{O}_2}$. For a Gaussian with mean $\boldsymbol{\mu}_j$ such that $j \in \mathbf{O}_2$, let $\boldsymbol{\mu}_i$ be the mean of the closest Gaussian in $\mathbf{O}_1$. Then, $\theta^{\boldsymbol{\mu}_j}_c$ is the angle between the vectors $\boldsymbol{\mu}_i-\mathbf{q}_2$ and $\boldsymbol{\mu}_i-\boldsymbol{\mu}_j$. When $\boldsymbol{\mu}_j$ is not intersecting $\mathbf{O}_1$, $\theta^{\boldsymbol{\mu}_j}_c<\pi/2$ (\cref{fig:physics_losses_supp}(b), left side), and when $\boldsymbol{\mu}_j$ is intersecting $\mathbf{O}_1$, $\theta^{\boldsymbol{\mu}_j}_c>\pi/2$ (\cref{fig:physics_losses_supp}(b), right side). To avoid intersection, we aim to enforce that the contact angle is acute for intersecting Gaussians. Then,
\begin{equation}
    \mathcal{L}_c=\frac{1}{\left|\mathcal{K}^{<\frac{\pi}{2}}_{\mathbf{O}_2}\right|}\underset{\boldsymbol{\mu}_j \in \mathcal{K}^{<\frac{\pi}{2}}_{\mathbf{O}_2}}{\sum}  -\cos \theta^{\boldsymbol{\mu}_j}_c,
\end{equation}
where $\mathcal{K}^{<\frac{\pi}{2}}_{\mathbf{O}_2}= \{\boldsymbol{\mu}_j \,\,|\,\, \boldsymbol{\mu}_j \in \mathcal{K}_{\mathbf{O}_2},\,\theta^{\boldsymbol{\mu}_j}_c>\pi/2\}$. 

The overall physics loss is therefore given by,
\begin{equation}
    \mathcal{L}_\textrm{Phys} = \lambda_g \mathcal{L}_g + \lambda_c \mathcal{L}_c,
\end{equation}
where $\lambda_g$ and $\lambda_c$ are appropriate regularization parameters. In practice we use $\lambda_g=10,000$. For $\lambda_c$, we find that $\lambda_c$ must balance the current gravity loss, to prevent the objects from remaining intersected. Therefore, in the case of object intersection, we set $\lambda_c=\mathcal{L}_g*30,000$. We find that 200 steps of the physics-guided optimization is sufficient to arrive at feasible compositions. These hyperparameters were found to work for our tested examples, but can be finetuned depending on specific use cases as well, based on the user.

\paragraph{Stabilizing impulse:} As mentioned in the main paper, in additiona to the above physics-based losses, we include a ``stabilizing impulse" as part of our composition. The necessity for this constraint arises out of limitations of 2D diffusion guidance for 3D composition. Specifically, since we use elevated camera perspectives, some projective ambiguities might creep through. Therefore, for prompts such as ``a hamburger on a plate", the hamburger might find itself not perfectly aligned with the plate and hence might overhang. Then, once the physical constraints are activated, this might lead to the hamburger falling off the plate, for example. Therefore, we incorporate the constraint as follows.

If the top-view cross section areas of $\mathbf{O}_1$ and $\mathbf{O}_2$ overlap in area between 40\% and 95\% of the cross section area of $\mathbf{O}_2$, and if $\mathbf{O}_1$ and $\mathbf{O}_2$ have come into contact, $\mathbf{O}_2$ receives a small impulse toward the central axis of $\mathbf{O}_1$, and upwards. This impulse is set to have a translation distance of 0.3 and an angle with the horizontal of 60 degrees. We limit this impulse to only act 5 times for a composition. This accounts for strong diffusion guidance against the impulse.

\section{Object Generation Details}
The primary objective function is the SDS loss that we minimize for $\textit{iterations}=3,000$, $\textit{batch size}=4$, $\textit{prompt}=y$  and $CFG=100$. We scale the loss by $0.5$, and linearly reduce the sampled timestep, $t$, interval from $[2,980]$ to $[2,500]$ for 2,000 steps for faster convergence. We rescale the SDS loss according to~\citet{lin2023common} by a factor of $0.7$ to reduce overexposure. We initialize the Gaussians using PointE's text-to-3D model~\cite{nichol2022point, chen2023text}. Given a text prompt, it returns 4,096 points that are converted to 3D isotropic Gaussians with random color features, $scale=0.02$, and $opacity=0.8$. We sample images of the 3D object from random views: azimuth from $[0^{\circ},360^{\circ}]$, elevation from $[-30^{\circ},80^{\circ}]$, and FOV from $[30^{\circ},55^{\circ}]$. The prompt $y$ is appended with text to describe the direction of the sampled object view according to~\citet{poole2022dreamfusion} to reduce Janus affects. The 3D Gaussians are updated according to the parameters defined in \cref{tab:3dgs_parameters}. 

\begin{table}[t]
    \centering
    \begin{tabular}{lc}
         \toprule
         Parameter & Value \\
         \midrule
         Iterations & 3000 \\
         Learning rate & \\
         \quad $\mu$ (Start, End)[exp. decay] & (2e-3,1e-4)\\
         \quad $c$& 0.01\\
         \quad $\sigma$& 0.01\\
         \quad $\Sigma$& 0.002\\
         Densification& \\
         \quad (Start, End) & (0, 2000)\\
         \quad Interval & 250 \\
         \quad Gradient Threshold& 0.5\\
         Max Spherical Harmonic& 0\\
         Opacity Reset Interval & $\emptyset$\\
    \end{tabular}
    \caption{\textbf{3D Gaussian Splatting hyperparameters for object generation.}}
    \label{tab:3dgs_parameters}
\end{table}

\subsection{Alpha Hull}
\begin{figure}[t]
  \centering
  \includegraphics[width=0.8\linewidth]{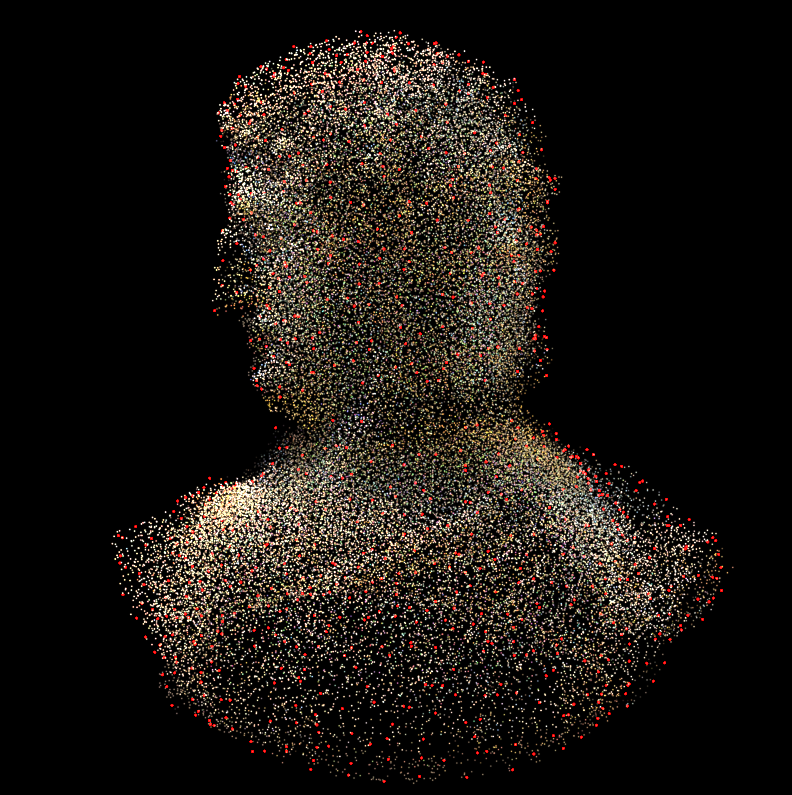}

   \caption{\textbf{We visualize the set of points belonging to the Alpha hull, $\mathcal{A}$.} For visibility, Gaussians have been scaled down to 1\%. The points belonging to $\mathcal{A}$ are marked in red, while the rest of the points keep their original color. The Alpha hull loss brings all Gaussians closer to the set $\mathcal{A}$. The original 3D object is in \cref{fig:alpha_ablation} in the right-most column.}
   \label{fig:alpha_shape}
\vspace{-4mm}
\end{figure}

\begin{figure*}[h]
    \centering
    \includegraphics[width=1\linewidth]{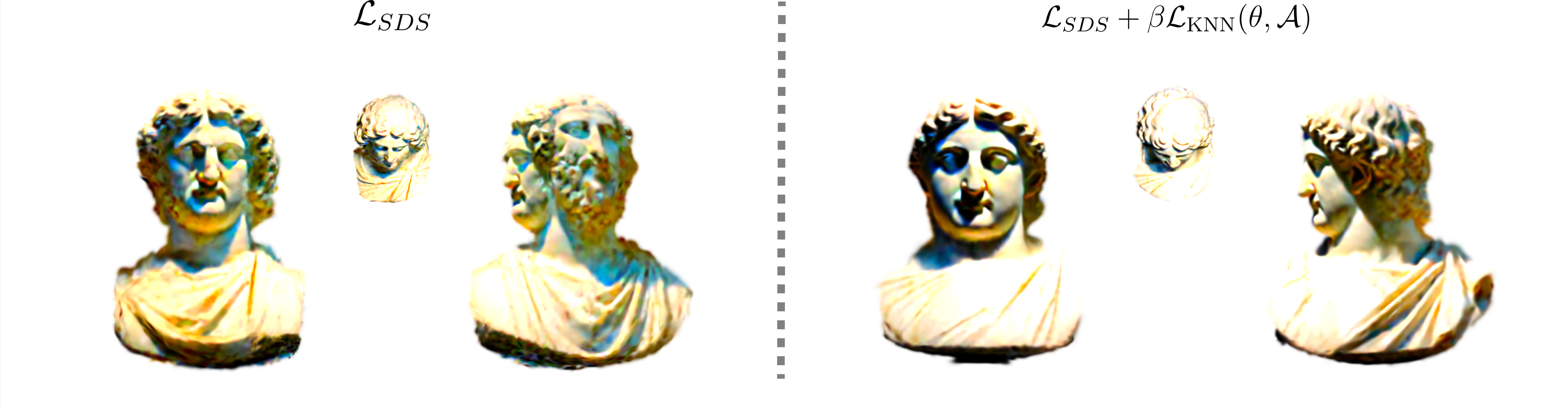}
    \caption{\textbf{Comparison between object generation without (left) and with (right) the Alpha hull proximity loss.} The object was generated with the prompt ``a Greek bust".}
    \label{fig:alpha_ablation}
\end{figure*}

The Alpha hull~\cite{Bellockk} loss, $ \beta\mathcal{L}_\text{KNN}(\theta,\mathcal{A})$, brings points to the surface to distribute Gaussians more uniformely and densely. We set $\beta=5$ and $K=5$. \cref{fig:alpha_shape} illustrates the set of points, $\mathcal{A}$, that are picked for an object. \cref{fig:alpha_ablation} shows a comparison between optimization with and without the Alpha hull loss. Without it, the surface of objects can appear blotchy and individual Gaussians are more visible. 

\begin{figure}[b]
  \centering
  \includegraphics[width=1\linewidth]{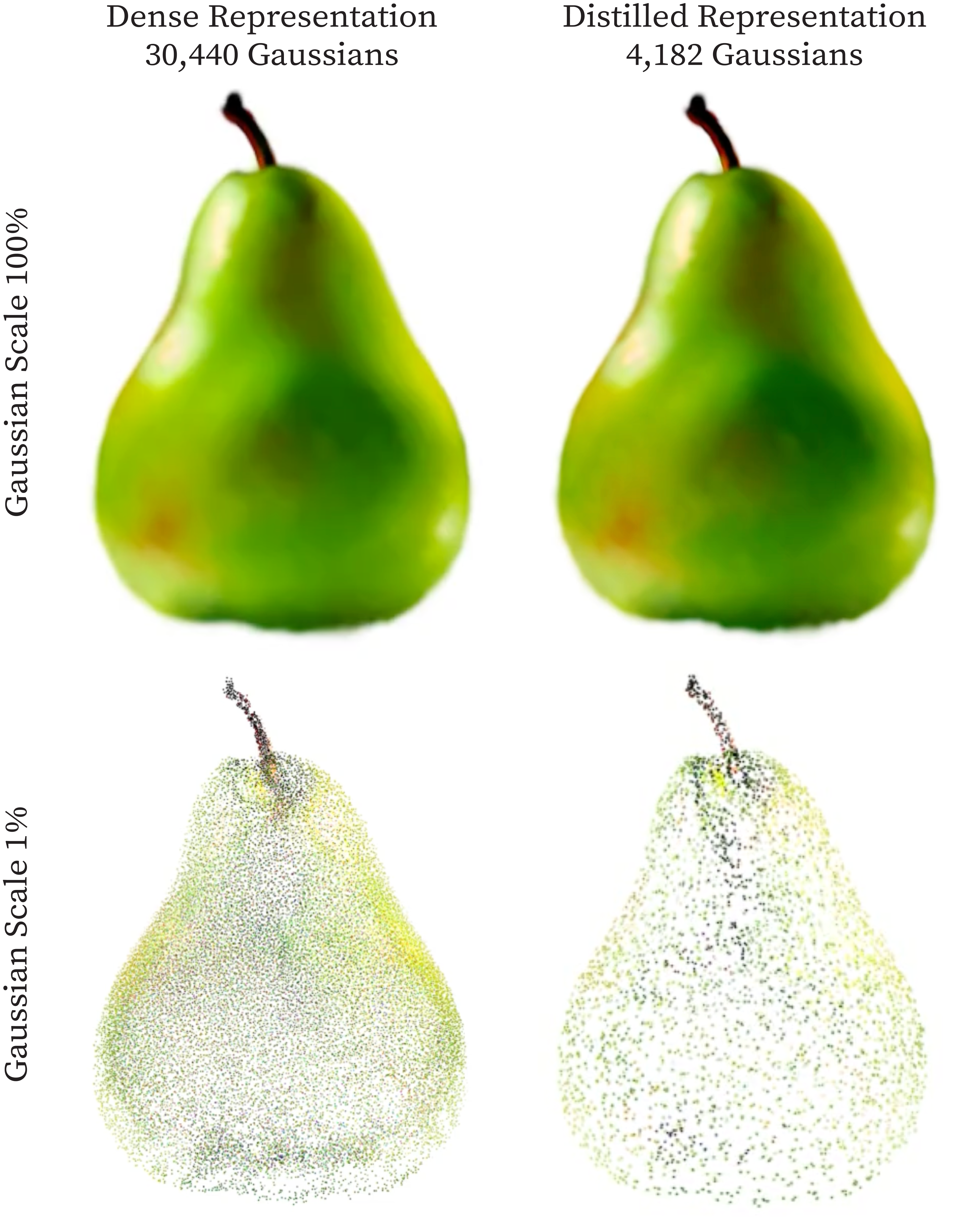}

   \caption{\textbf{Visualization of object distillation for a ``pear".} }
   \label{fig:distill}
\vspace{-4mm}
\end{figure}

 \begin{figure*}[h]
    \centering
    \includegraphics[width=1\linewidth]{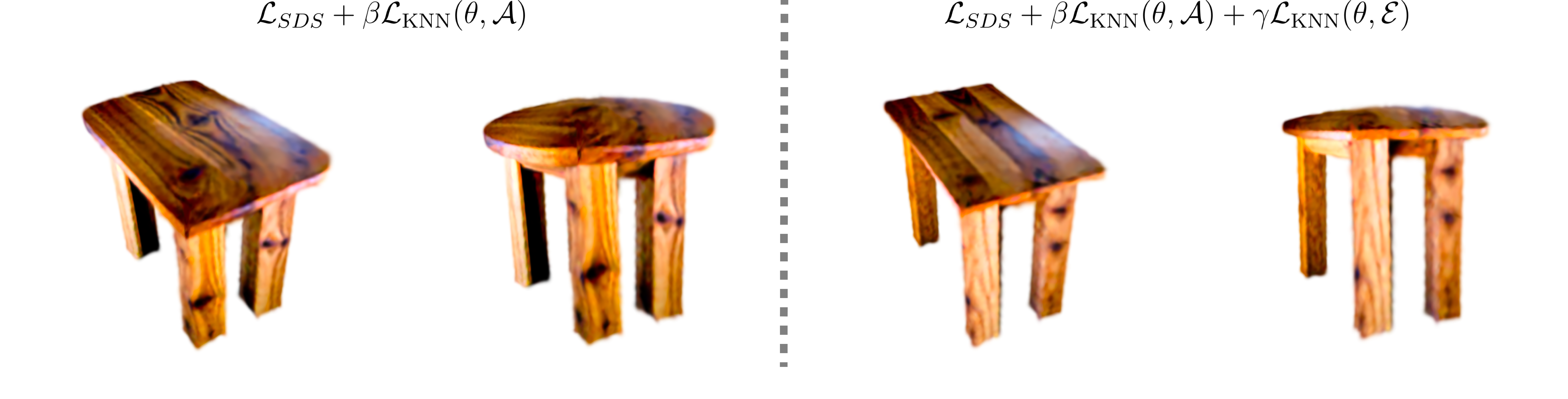}
    \caption{\textbf{Comparison between object generation without (left) and with (right) the PointE initialization proximity loss.} The object was generated with the prompt ``a wooden table".}
    \label{fig:pointe_ablation}
\end{figure*}

\subsection{PointE KNN}
The optional PointE KNN provides crucial geometrical guidance to enable generation of certain objects with geometries that have 3D ambiguities under image projection. The PointE KNN constrains Gaussians to be near the set of initialization points provided, $\mathcal{E}$. Then, the KNN loss is $\gamma\mathcal{L}_\text{KNN}(\theta,\mathcal{E})$ (defined in \cref{eq:knn}), where $\gamma=20$ and $K=1$. We find this loss to be instrumental in two cases: preserving flat surfaces and maintaining convexity. While the loss is not strictly required for a single object to appear accurate under 2D projection, its absence can have consequences for 3D composition. For example, we find that objects that have concavities, such as a basket, open box, or cup, will have the concavity filled or partially filled during optimization without the geometrical regularization. This results in failed compositions where another object is supposed to fill the concavity. Similarly, without the PointE KNN, a flat surface will begin to become curved and convex. This is apparent in \cref{fig:pointe_ablation}. In column 3, we can see that the top view appears reasonable, however, the side view shows the convexity of the table's surface. Whereas with the PointE KNN, in column 4, such effects are reduced. For composition, flat surfaces are necessary to preserve physically realistic scenes, especially from side views. On the other hand, PointE KNN can be detrimental if applied to all objects. Strict geometrical regularization to the initialized shape can cause object's to not develop new features. For this reason, the PointE KNN is optional and should be supplied as input by a user. 

\subsection{Memory Distillation}
 The geometry of the generated objects is not well-defined during training, unlike in the novel-view synthesis setting with well-defined supervision. Object generation with the 3D Gaussian splatting framework requires frequent duplication of Gaussians to enable high-frequency details and expressability of the model. Therefore, due to overzealous duplication, redundant Gaussians are unavoidable. Moreover, for large-scale scene generation with numerous objects, we need to avoid high memory usage. This problem can be addressed by distilling the dense representation of a generated object. This is done by retraining a new 3D Gaussian model on views from the original dense object. In fact, this amounts to reducing the problem to classic novel-view synthesis. As a bonus, this may be even more stable due to the availability of an infinite number of views of the object from various camera positions and radii. Quantitative results are shown in \cref{tab:distillation} and a qualitative result is shown in \cref{fig:distill} where the scale of the Gaussians has been reduced to show the effect of distillation.

 \section{User Input Details}
 As illustrated in \cref{fig:pipeline}, the composition-level text prompt can be explicitly decomposed into the object-level prompts and interaction prompts. The user then has to provide two forms of input: object-level and interaction text prompts. Formally, for each object $\textbf{O}_i$, there must be a corresponding text prompt describing the object $\textbf{y}_i$. Similarly, for all pairs of objects with interactions $\textbf{P}_{i,j}$, there must also be a corresponding textual description of the interaction $\textbf{y}_{i,j}$. The interaction text prompt must reference the objects participating in the interaction, $(\textbf{O}_i,\textbf{O}_j)$, but does not necessarily need to include the object's original text attributes such as color or style. For example, given the composition-level text prompt, ``a roasted chicken on a plain white plate laying on a wooden table", it can be decomposed as follows: 
 \begin{enumerate}[leftmargin=30pt]
     \item Objects:
     \begin{enumerate}
         \item $\textbf{O}_1$ = \textit{photo of a roasted chicken}
         \item $\textbf{O}_2$ = \textit{photo of a plain white plate}
         \item $\textbf{O}_3$ = \textit{photo of a wooden table}
     \end{enumerate}
     \item Interactions:
     \begin{enumerate}
         \item $\textbf{P}_{1,2}$ = \textit{roasted chicken on a plate}
         \item $\textbf{P}_{2,3}$ = \textit{a plate on a table}
     \end{enumerate}
 \end{enumerate}
 which can then be supplied as input to our model.

\begin{figure*}[h]
    \centering
    \includegraphics[width=1\linewidth]{figures_supp/f_big_results/supplement_big_results_A.png}
    \caption{\textbf{Additional Composition Results 1.}}
    \label{fig:A-label}
\end{figure*}

\begin{figure*}[h]
    \centering
    \includegraphics[width=1\linewidth]{figures_supp/f_big_results/supplement_big_results_B.png}
    \caption{\textbf{Additional Composition Results 2.}}
    \label{fig:B-label}
\end{figure*}

\section{Additional Results}
We include additional results for various compositions of objects in \cref{fig:A-label} and \cref{fig:B-label} on the following pages.